\documentclass[11pt]{article}

\usepackage[final]{acl}

\usepackage{times}
\usepackage{latexsym}

\usepackage[T1]{fontenc}

\usepackage[utf8]{inputenc}

\usepackage{microtype}

\usepackage{inconsolata}

\usepackage{graphicx}
\usepackage{booktabs}
\usepackage{xcolor}
\usepackage{tabularx}  
\usepackage{multirow}
\usepackage{tikz}
\usepackage{csvsimple}
\usepackage{pgfplots}
\usepackage{pgfplotstable}
\usepackage{subcaption}
\usepackage[table]{xcolor}
\pgfplotsset{compat=1.18}
\usepackage{amssymb}

\definecolor{RTOne}{HTML}{E0B12B}        
\definecolor{SCOUT}{HTML}{EF6650}        
\definecolor{ALFRED}{HTML}{7268D8}       
\definecolor{BRIDGE}{HTML}{3185FF}       
\definecolor{TacoPlay}{HTML}{4DB6AC}     
\definecolor{LanguageTable}{HTML}{8BD47E}
\definecolor{LIBERO}{HTML}{E94E77}       
\title{Limited Linguistic Diversity in Embodied AI Datasets} 

\author{
  \textbf{Selma Wanna\textsuperscript{$\dagger$ 1}},
  \textbf{Agnes Luhtaru\textsuperscript{$\dagger$ 2}},
  \textbf{Jonathan Salfity\textsuperscript{3}},
  \textbf{Ryan Barron\textsuperscript{1}},
\\
  \textbf{Juston Moore\textsuperscript{1}},
  \textbf{Cynthia Matuszek\textsuperscript{4}},
  \textbf{Mitch Pryor\textsuperscript{3}}
\\
\\
  \textsuperscript{1}Los Alamos National Laboratory, Los Alamos, USA, \\
  \textsuperscript{2}Institute of Computer Science, University of Tartu, Estonia, \\
  \textsuperscript{3}Department of Mechanical Engineering, The University of Texas at Austin, USA, \\
  \textsuperscript{4}University of Maryland, Baltimore County, Maryland, USA\\
  \small{
     \textbf{Equal contribution:} \textsuperscript{$\dagger$} denotes equal contribution; \textbf{Correspondence:} \href{mailto:slwanna@lanl.gov}{slwanna@lanl.gov}, \href{mailto:agnes.luhtaru@ut.ee}{agnes.luhtaru@ut.ee}
  }  \\ 
}

\begin{document}
\maketitle
\begin{abstract}
Language plays a critical role in Vision-Language-Action (VLA) models, yet the linguistic characteristics of the datasets used to train and evaluate these systems remain poorly documented. In this work, we present a systematic dataset audit of several widely used VLA corpora, aiming to characterize what kinds of instructions these datasets actually contain and how much linguistic variety they provide. We quantify instruction language along complementary dimensions—including lexical variety, duplication and overlap, semantic similarity, and syntactic complexity. Our analysis shows that many datasets rely on highly repetitive, template-like commands with limited structural variation, yielding a narrow distribution of instruction forms. We position these findings as descriptive documentation of the language signal available in current VLA training and evaluation data, intended to support more detailed dataset reporting, more principled dataset selection, and targeted curation or augmentation strategies that broaden language coverage.
\end{abstract}

\section{Introduction}
\label{sec:intro}
With advances in large language models (LLMs) and multimodal learning, language is increasingly used as an input across research fields, enabling practical systems. In robotics, this is reflected in the growing focus on Vision-Language-Action (VLA) models such as OpenVLA \cite{pmlr-v270-kim25c}, RT-X \cite{openxembodiment}, and $\pi$0.5 \cite{intelligence2025pi05visionlanguageactionmodelopenworld}. Much of this progress has been driven by datasets like Open X-Embodiment \cite[OXE,][]{openxembodiment}, which are larger and more diverse across objects, scenes, and embodiments than earlier robotics datasets, supporting a shift toward end-to-end generalist robotic systems that use language to specify tasks.
\begin{figure}[t]
\small
\centering

\newcommand{\verbcol}[1]{\textcolor{ALFRED!85!black!95}{#1}}
\newcommand{\objcol}[1]{\textcolor{LIBERO!85!black!95}{#1}}
\newcommand{\adpcol}[1]{\textcolor{TacoPlay!85!black!95}{#1}}
\newcommand{\adjcol}[1]{\textcolor{RTOne!85!black!95}{#1}}

{\small\textbf{RT-1}} \\[0.9ex]

{\color{black!25}\textbf{\verbcol{move} \objcol{sponge} \adpcol{near} \adjcol{orange} \objcol{can} \\[0.5ex]
\verbcol{move} \adjcol{blue plastic} \objcol{bottle} \adpcol{near} \objcol{sponge}
}}

\vspace{0.4ex}\rule{0.9\linewidth}{0.3pt}\vspace{0.4ex}

{\small\textbf{BridgeData}} \\[0.9ex]

{\color{black!55}\textbf{
\verbcol{pick} the \objcol{carrot} and \verbcol{place} it \adpcol{inside} the \adjcol{stainless steel} \objcol{pot} \\[0.5ex]
\verbcol{pick} the \objcol{strawberry} and \verbcol{put} it \adpcol{into} the \adjcol{stainless} \objcol{pot}.
}}

\vspace{0.4ex}\rule{0.9\linewidth}{0.3pt}\vspace{0.4ex}

{\small\textbf{Language Table}} \\ 
\vspace{0.9ex}

{\color{black!55}\textbf{
\verbcol{slide} the \adjcol{red} \objcol{star} \adpcol{into} the \objcol{cube} \\[0.5ex]
slightly \verbcol{move} the \adjcol{red} \objcol{circle} \adpcol{below} the \adjcol{red} \objcol{star}
}}

\caption{Examples of language instructions from selected VLA datasets in the OXE collection, illustrating limited linguistic diversity. Colored tokens denote verbs (\verbcol{purple}), object nouns (\objcol{pink}), spatial relations (\adpcol{teal}), and descriptive adjectives (\adjcol{yellow})}
\label{fig:compact_commands}
\end{figure}

Despite this progress, language---while a core modality in VLA systems---is often overlooked in dataset documentation and evaluations. Datasets in the OXE collection were not originally created for generalist VLA training, and documentation emphasizes diversity across objects, scenes, and embodiments, with little discussion of instruction language \cite{openxembodiment}. Meanwhile, a growing body of work reports limited semantic robustness of VLA models despite reliance on LLM backbones, including sensitivity to paraphrases, performance drops in the presence of distractor objects, and related failures \cite{gao2025taxonomyevaluatinggeneralistrobot, contributors2025agibotworld, wang2024ladevlanguagedriventestingevaluation}. These generalization issues suggest that language understanding may be underemphasized in current VLA development; documenting the linguistic features of the data that shape model performance is a crucial step for future research.

\begin{figure*}[t]
    \centering
    \includegraphics[width=\linewidth]{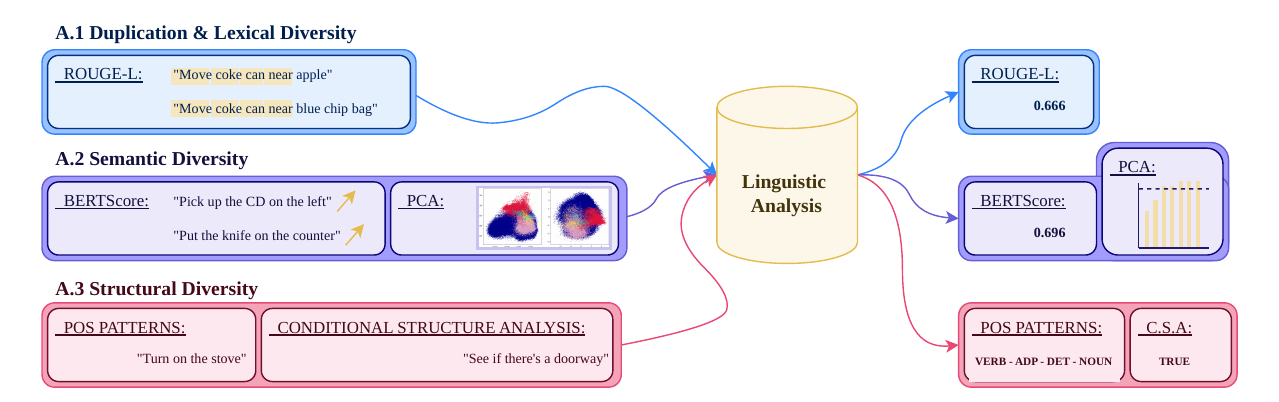}
    \caption{We analyze linguistic diversity in embodied AI datasets across three categories: Analysis 1 (A.1) Duplication \& Lexical diversity, Analysis 2 (A.2) Semantic diversity, and Analysis 3 (A.3) Structural diversity.}
    \label{fig:feat-pipeline}
\end{figure*}

To address this, we analyze instruction language in several VLA datasets from OXE and compare it to other robotics datasets and common instruction-tuning corpora used for LLM training. We characterize language coverage across three dimensions: lexical redundancy, semantic diversity, and structural diversity (cf. Section \ref{sec:methodology}). We find that current VLA datasets exhibit a narrow and repetitive instruction distribution: fewer than 2\% of instructions are unique, lexical diversity is limited relative to the comparison datasets, and structural forms are often templated (cf. Section \ref{sec:results}). Beyond repetition, linguistically richer constructions are largely missing: across all studied datasets, fewer than 1\% of commands contain negation, and conditionals are similarly rare.

Overall, we provide a systematic characterization of language used in VLA datasets (see Figure~\ref{fig:compact_commands} for command examples). By making these properties explicit, this language-focused documentation can inform synthetic data generation and future data collection, help contextualize reported generalization gaps such as sensitivity to paraphrases, and guide the design of new evaluation protocols.

\section{Background}
\label{sec:background}
Many recent VLA models build on modern vision--language models (VLMs) that combine a vision encoder and an LLM via adapters (e.g., PaliGemma) \cite{beyer2024paligemmaversatile3bvlm}. Action generation is typically added either autoregressively, by extending the LLM vocabulary with action tokens \cite{kim2024openvla, pertsch2025fastefficientactiontokenization}, or via a separate action expert trained with diffusion/flow-matching objectives \cite{black2024pi0visionlanguageactionflowmodel}.

A key challenge is integrating action supervision without catastrophic forgetting or degrading pretrained VLM capabilities, which lead to semantic generalization failures. Prior work studies how action representations affect language generalization \cite{grover2025enhancinggeneralizationvisionlanguageactionmodels} and how combining an autoregressive VLM with an action expert can change or degrade VLM behavior \cite{driess2025knowledge}. Other works investigate co-training with multimodal understanding data \cite{zhou2025chatvlaunifiedmultimodalunderstanding, gao2025taxonomyevaluatinggeneralistrobot, grover2025enhancinggeneralizationvisionlanguageactionmodels}, or incorporate reasoning-style supervision, resulting in slower inference \cite{zawalski2024robotic, chen2025trainingstrategiesefficientembodied}.

In contrast, there is comparatively limited work on what training-data properties support effective VLA grounding. Prior results in vision--language grounding suggest that coverage and data quality are critical for alignment, and that gaps in coverage are associated with failures even on simple capabilities, with performance degrading on underrepresented concepts and improving with targeted coverage \cite{udandarao2024no, zhang2024why}. For VLA models, however, dataset analyses have mostly focused on non-linguistic sources of shortcuts (e.g., fragmentation, viewpoints, backgrounds); \citet{xing2025shortcut} highlights such issues and briefly notes limited language variation, but does not provide a detailed characterization of instruction language.

While these findings do not isolate linguistic diversity as a causal factor in VLA performance, they suggest that gaps in training data coverage can limit generalization and language–action grounding even when the underlying backbone has strong capabilities, motivating the need to study linguistic variability in VLA datasets.

\section{Measuring Linguistic Diversity for EAI}
\label{sec:methodology}
There is no widely adopted standard for characterizing linguistic diversity in embodied AI (EAI) datasets. Dataset \emph{diversity} is inherently vague, and many studies claim to address diversity without clearly defining what it entails \cite{pmlr-v235-zhao24a}. Linguistic diversity itself spans multiple dimensions: for example, \citet{guo-etal-2024-curious} discusses lexical diversity (variation in word choice), semantic diversity (variation in meaning), and syntactic diversity (variation in grammatical structure). \citet{tevet-berant-2021-evaluating} distinguishes between form diversity, which relates to surface-level variation, and content diversity, which captures deeper semantic differences.

In VLA datasets, however, diversity is typically defined in non-linguistic terms, such as the number of distinct objects, environments, or tasks \cite{openxembodiment}. While this framing may reflect aspects of semantic variation (e.g., which actions are performed on which objects), it provides little insight into the diversity of natural-language instructions, including variation in word choice, grammatical structure, and meaning expression.

We construct a diversity evaluation that incorporates a broad range of analysis metrics, largely following the categorization followed by \citet{tevet-berant-2021-evaluating}. Our evaluation covers three analyses (see Figure~\ref{fig:feat-pipeline}): 
\begin{itemize}
    \item \textbf{Analysis 1: Duplication and Lexical Diversity}, including standard statistics (e.g., number of unique unigrams, sentence length, lexical overlap) and common diversity metrics;
    \item \textbf{Analysis 2: Semantic Diversity}, measured using embedding based methods, and verb–direct object–adverbial diversity;  
    \item \textbf{Analysis 3: Structural Diversity}, including syntax (POS-patterns) and higher-level linguistic phenomena, like negations.
\end{itemize}

 Each analysis targets a distinct aspect of linguistic diversity, enabling a more comprehensive understanding of the textual variation present in datasets used to train models. In this work, we focus specifically on the repetitiveness and uniformity of language in VLA datasets, rather than on diversity related to sociocultural or demographic bias. Below, we describe each analysis in detail. The corresponding results are presented in Section \ref{sec:results}. We discuss additional qualitative patterns in selected datasets in the Appendix~\ref{sec:qual-feat}.

\subsection{Duplication and Lexical Diversity}

Many OXE datasets were created primarily for imitation learning, where language is often treated as an auxiliary signal. Consequently, datasets built from human-teleoperated or scripted demonstrations frequently exhibit substantial repetition in instruction phrasing and limited coverage of distinct command types. In the LLM literature, duplication is associated with increased memorization and reduced reliance on generalization \cite{kandpal2022deduplicating, lee-etal-2022-deduplicating}. More broadly, deep networks have sufficient capacity to fit random labels, highlighting that memorization can be easy in overparameterized settings \cite{zhang2017understanding}. We hypothesize that similar risks may arise for VLA models trained on highly repetitive instructions.

In discussions of linguistic diversity—especially in the context of LLM-generated text—lexical diversity is typically the primary focus. Metrics such as BLEU \cite{papineni-etal-2002-bleu, selfbleu} and ROUGE \cite{lin-2004-rouge}, although originally designed for evaluation of generation quality, have been adapted to approximate diversity. Similarly, compression ratio (CR), based on gzip, has been used as a proxy for textual diversity and has been shown to be effective at distinguishing LLM-generated from human-authored text \cite{shaib2025standardizingmeasurementtextdiversity}. 

To assess duplication and lexical diversity, we compute basic linguistic statistics—such as the number of unique commands, sentences, and unigrams—as well as five textual diversity metrics.  ROUGE-L and compression ratio (CR), following \citet{shaib2025standardizingmeasurementtextdiversity}, are reported in Table \ref{tab:combined} while the remaining three metrics: BLEU, Jaccard similarity and Levenshtein distance are provided in Table \ref{tab-text-sim-ext}.

\subsection{Semantic Diversity}

Semantic diversity is typically measured using embedding-based metrics that group paraphrases closely in semantic space. While sentence embeddings such as BERTScore \cite{bert-score} often struggle to capture aspects like syntax, antonymy, and word order \cite{zhang-etal-2023-well, mahajan-etal-2024-align}, they can be used for evaluating variation in content (i.e., what is said, not how) \cite{tevet-berant-2021-evaluating, stasaski-hearst-2022-semantic}. In the context of robotics, such metrics can help quantify the variety of actions and object interactions described in natural language instructions. Among the dimensions we evaluate, semantic diversity is expected to correlate most closely with the number of distinct objects and skills in the dataset, as these reflect the variety of task meanings that natural language instructions encode. 

Motivated by this, we include BERTScore as a pairwise diversity metric between individual instructions. In addition, we apply Principal Component Analysis (PCA) to sentence embeddings of entire datasets and report the number of components required to explain 95\% of the cumulative variance \cite{pca-intrinsic-dim, intrinsic-dim-non-lin-pca}, to capture dataset-level variation in instruction semantics. For sentence embeddings, we use four common encoders - USE (512D)~\cite{cer-etal-2018-universal}, SBERT (768D)~\cite{reimers-gurevych-2019-sentence}, CLIP (512D)~\cite{CLIP}, and SONAR (1024D)~\cite{Duquenne:2023:sonar_arxiv}. For brevity, we report USE in Table \ref{tab:combined} but provide the results across all encoders in Table \ref{tab-pca-full}. We further justify this approach in the Appendix~\ref{sec:pca-extended}. 

In addition to these metrics, we examine verb, direct object, and adverb diversity to provide more interpretable and domain-specific insights for robotics. Specifically, we assess how many distinct verbs are used with each direct object in manipulation datasets. In contrast, direct object structures are less relevant for navigation-focused datasets. For these, the manner in which an instruction is followed—such as the use of directional terms (e.g., “north,” “forward”), location-based modifiers (e.g., “around,” “inside”), and manner descriptors (e.g., “slowly,” “directly”)—is more pertinent. Low counts of these features suggest limited interaction diversity, which could introduce biases. If a model has learned to perform only specific actions with certain objects, it may eventually learn to ignore the language command, as neural networks have a well-known simplicity bias \cite{10.5555/3495724.3496527}.

\subsection{Structural Diversity}

We group both syntactic variation and the presence of higher-level linguistic phenomena—such as negation, conditionals, cycles, and multi-step instructions—under the category of structural diversity. Prior work has shown that limited syntactic variety can amplify model biases, whereas incorporating more diverse grammatical structures improves generalization and reduces overfitting to shallow heuristics \cite{aggarwal-etal-2022-towards}. Common measures of syntactic diversity include part-of-speech (POS) patterns and constituency or dependency parse trees. In our evaluation, we include the frequency of different POS patterns and use constituency tree similarity \cite{moschitti-2006-making} to quantify syntactic variation across datasets (reported in Table \ref{tab:combined}).

Beyond surface-level syntax, structural diversity also includes compositional and logic-oriented constructions. Logical structures can support reasoning capabilities in language models \cite{uchiyama-etal-2024-programming}, and challenges in handling negation remain prominent in NLU tasks \cite{hossain-etal-2022-analysis} and \citet{zhang-etal-2025-negvqa} demonstrate that VLMs struggle significantly with negation. In the context of robotics, the presence of linguistic phenomena such as negation, conditionals, and multi-step instructions is particularly important. Even if these structures do not strongly affect current VLA benchmark scores, they remain important because they express constraints, exceptions, contingency, and sequential logic that arise naturally in real-world interaction. These structures increase the complexity of commands that a robot must understand and execute. Without them, instructions are limited to simple, atomic actions, reducing the system’s ability to interpret nuanced or context-dependent behaviors. For example, instructions such as \textit{``give me the apple that is not rotten''}, \textit{``if you have picked up the apple, wash it''}, or \textit{``pick up the apple and place it on the cutting board''} require compositional understanding and the ability to process conditional logic, negation, and sequential actions. Capturing such structures in training data may also support reasoning and help models handle more complex commands.

To quantify this aspect of structural diversity, we estimate the proportion of instructions that contain negation, conditionals, cycles, or multi-step commands. We rely on syntactic cues from dependency parses, specific keyword patterns, and part-of-speech tags to identify these phenomena. In addition, we manually annotate a subset of each dataset to validate the accuracy of the automatic heuristics and to better characterize the linguistic structures present.

\begin{table*}[ht!]

\begin{center}
\scriptsize
\begin{tabular}{lccccc}
\toprule
\textbf{Dataset} & \textbf{Citations} & \textbf{Focus} & \textbf{Language Style} \\
\midrule

\multicolumn{4}{l}{\textit{Instruction-Tuning }} \\
\hspace{3mm}OASST2~\cite{2023openassistant} & 779+ & LLM instruction tuning & Crowdsourced instructions \\
\hspace{3mm}Alpaca~\cite{alpaca} & 4361+ & LLM instruction tuning & Generated from seed data \\
\hspace{3mm}LLaVA-Instruct~\cite{NEURIPS2023_6dcf277e} & 7286+ & VLM instruction tuning & Generated, specific questions \\

\midrule
\multicolumn{3}{l}{\textit{Language-Focused Robotics Datasets}} \\
\hspace{3mm}ALFRED~\cite{Shridhar_2020_CVPR} & 936+ & Household task instruction following & Step-by-step, high-level \\
\hspace{3mm}SCOUT~\cite{lukin-etal-2024-scout-situated} & 6+ &  Two-way, task-oriented dialogue & Unconstrained, interactive \\
\midrule
\multicolumn{3}{l}{\textit{VLA Datasets}} \\
\hspace{3mm}Open X-Embodiment ~\cite{openxembodiment} & 746+ & Collection of datasets & Varied, not always included \\
\hspace{6mm}RT-1~\cite{brohan2023rt1roboticstransformerrealworld} & 1693+ & Kitchen instruction following & Concise, imperative, templated \\
\hspace{6mm}BRIDGE~\cite{bridge_2022} & 330+ & Skill generalization across domains & Diverse, step-by-step \\
\hspace{6mm}TacoPlay~\cite{rosete2022tacorl} & 108+ &  Task-agnostic ``play” behaviors & Simple, low-variety, templated \\
\hspace{6mm}Language Table~\cite{langtable} & 290+ &  Open-vocab spatial manipulation & Natural, open-ended \\
\hspace{3mm}LIBERO~\cite{liu2023liberobenchmarkingknowledgetransfer} & 450+ & Knowledge transfer in robot learning & Natural\footnote{The LIBERO-10 commands are taken from Ego4D \cite{Grauman_2022_CVPR} then used to develop language templates.} \\

\bottomrule
\end{tabular}
\caption{Overview of the datasets explored in this work. We include citation counts for each dataset; note that some of the referenced works focus primarily on dataset creation, while others introduce new methods alongside the dataset. Citation data from Semantic Scholar (Alpaca from Google Scholar).}
\label{tab-eai-datasets-overview}
\end{center}
\end{table*}

\section{Datasets}
\label{sec:datasets}
To investigate the linguistic characteristics of data used to train VLA models, we analyze a set of datasets drawn primarily from the OXE collection, which are among the most widely used sources for VLA training \cite{kim2024openvla, openxembodiment, intelligence2025pi05visionlanguageactionmodelopenworld}. OXE contains over 40 datasets with language annotations; we therefore focus on a representative subset chosen to balance (i) relevance/usage in the literature (e.g., citation frequency and common inclusion in training or selection pipelines), (ii) scale (sufficient episodes for stable estimates), and (iii) coverage of distinct linguistic regimes (templated vs. natural, open-ended instructions). An overview is provided in Table~\ref{tab-eai-datasets-overview}, with further details in Section~\ref{sec:vla-datasets}.

As many of the computed metrics are difficult to interpret in isolation, we include additional reference datasets to contextualize the results: robotics datasets that are not typically used to train generalist VLA models but place greater emphasis on language, and (in selected experiments) instruction-tuning  datasets from outside robotics. These references are not intended to define “ideal” properties; rather, they serve as anchors for interpreting the linguistic characteristics of existing VLA training corpora. In non-robotics instruction-following datasets, "commands" refer to individual sentence examples.

\subsection{VLA Datasets}
\label{sec:vla-datasets}

From the OXE collection, we analyze four widely studied datasets—RT-1 \cite{brohan2023rt1roboticstransformerrealworld}, BRIDGE \cite{bridge_2022}, TacoPlay \cite{rosete2022tacorl}, and Language Table \cite{langtable}. We include RT-1 \cite{brohan2023rt1roboticstransformerrealworld} because it is frequently cited, was introduced alongside the model, is among the largest OXE datasets in terms of episodes, and is also often selected by automatic data selection methods \cite{pmlr-v270-hejna25a, dass2025datamilselectingdatarobot}; its instructions are primarily imperative and highly templated. To contrast this with more natural language, we include Language Table \cite{langtable}, which has the most episodes in OXE and targets open-vocabulary spatial manipulation in controlled tabletop environments with more open-ended, dialogue-like interactions. We further include BRIDGE \cite{bridge_2022} as a complementary generalization-oriented dataset: while it also aims to support broad task generalization, it covers a wider range of tasks and environments, supports tool use and fine-grained object interactions, and is commonly featured in generalization evaluations.\footnote{We use the original Bridge dataset from the OXE download link. A newer version, which is now more commonly used, contains more episodes \cite{walke2023bridgedata}.} Finally, we include TacoPlay \cite{rosete2022tacorl} to represent a task-agnostic “play” paradigm learned from unstructured interaction data; although its language remains largely templated.

We additionally include LIBERO \cite{liu2023liberobenchmarkingknowledgetransfer}, a simulation-based benchmark focused on knowledge transfer. While not part of the OXE collection, LIBERO has been increasingly adopted in recent experimental and evaluation setups.

\subsection{Language-Focused Robotics Datasets}

We also include two robotics datasets that prioritize language interaction, although they are not primarily used to train generalist VLA models. ALFRED \cite{Shridhar_2020_CVPR} emphasizes natural language through fine-grained, step-by-step instructions aligned with low-level actions, making it well-suited for studying task decomposition and instruction-following.

SCOUT \cite{lukin-etal-2024-scout-situated} contains the most spontaneous, dialogue-based interaction data among the datasets considered. It captures unconstrained human-robot communication in navigation scenarios, supporting adaptive, context-aware exchanges beyond static command formats. SCOUT includes transcriptions from real-world robot operators and provides detailed linguistic statistics. For our analysis, we focus on utterances from the robot commander role to align with the style of other datasets.

\begin{table*}[ht!]
\begin{center}
\scriptsize
\setlength{\tabcolsep}{1.4mm}
\begin{tabular}{lccccccccc}
\toprule
Dataset & & \multicolumn{4}{c}{A.1 Duplication \& Lexical} & \multicolumn{2}{c}{A.2 Semantic} & A.3 Structural \\
\addlinespace[0.25em]
\cmidrule(lr){3-6}\cmidrule(lr){7-8}\cmidrule(lr){9-9}
\addlinespace[0.25em]
 & \# Sent & \# Uniq (\% Uniq) & \# Words & CR $\downarrow$ & ROUGE-L $\downarrow$ & BERTScore $\downarrow$ & USE $\uparrow$ & Tree Kernel $\downarrow$ \\
\midrule
\multicolumn{9}{l}{\textit{Instruction-Tuning Datasets}} \\
\hspace{3mm}OASST2~\cite{2023openassistant} & 42K+ & 39,301 (93.33\%) & \textbf{35,445} & \textbf{2.75} & \textbf{0.05 $\pm$  0.00} & \textbf{0.45 $\pm$ 0.00} & \textbf{254} & \textbf{2.25 $\pm$ 0.01 \%} \\
\hspace{3mm}Alpaca~\cite{alpaca} & 53K+ & 52,996 (\textbf{99.81\%}) & 18,141 & 3.20 & 0.10 $\pm$ 0.00 & 0.57 $\pm$ 0.00 & 231 & 3.66 $\pm$ 0.05 \% \\
\hspace{3mm}LLaVA-Instruct~\cite{NEURIPS2023_6dcf277e} & \textbf{366K+} & \textbf{261,892} (71.45\%) & 15,477 & 4.41 & 0.21 $\pm$ 0.00 & 0.61 $\pm$ 0.00 & 184 & 7.46 $\pm$ 0.13 \% \\
\midrule
\multicolumn{9}{l}{\textit{Language-Focused Robotics Datasets}} \\
\hspace{3mm}ALFRED~\cite{Shridhar_2020_CVPR} & \textbf{162K+} &  \textbf{126,005} (\textbf{79.9\%}) & \textbf{2,627} & 5.91 & 0.21 $\pm$ 0.00 & 0.64 $\pm$ 0.00 & \textbf{159} & 5.71 $\pm$ 0.14 \% \\
\hspace{3mm}SCOUT~\cite{lukin-etal-2024-scout-situated} &  23K+ & 8,795 (39.4\%) & 1,631 & \textbf{4.85} & \textbf{0.07 $\pm$ 0.00} & \textbf{0.49 $\pm$ 0.00} & 148 & \textbf{1.89 $\pm$ 0.22} \% \\
\midrule
\multicolumn{9}{l}{\textit{VLA Datasets}} \\
\hspace{3mm}RT-1~\cite{brohan2023rt1roboticstransformerrealworld} & 3.7M+ & 577 (0.02\%) & 49 & 118.20 & 0.19 $\pm$ 0.01 & 0.64 $\pm$ 0.00 & 33 &  5.09 $\pm$ 0.20 \% \\
\hspace{3mm}BRIDGE~\cite{bridge_2022} & 864K+ & 11,693 (1.4\%) &\textbf{1,189} & 64.90 & \textbf{0.15 $\pm$ 0.00} & \textbf{0.60 $\pm$ 0.00} & \textbf{125} & \textbf{3.68 $\pm$ 0.12 \%} \\
\hspace{3mm}TacoPlay~\cite{rosete2022tacorl} & ~214K &  403  (0.2\%) & 74 & 158.86 & 0.30 $\pm$  0.01 & 0.68 $\pm$ 0.00 & 42 & 8.86 $\pm$ 0.13 \% \\
\hspace{3mm}LanguageTable~\cite{langtable} & \textbf{7.0M+} &  \textbf{127,370} (\textbf{1.81\%}) & 928 & \textbf{56.64} & 0.29 $\pm$ 0.00 & 0.70 $\pm$ 0.00 & 86 & 9.19 $\pm$ 0.14 \% \\
\hspace{3mm}LIBERO~\cite{liu2023liberobenchmarkingknowledgetransfer} &  6.5K & 112 (1.72\%) & 79 & 134.86 & 0.38 $\pm$ 0.00 & 0.71 $\pm$ 0.00 & 34 & 12.22 $\pm$ 0.29 \% \\
\bottomrule
\end{tabular}
\end{center}
\caption{Summary of all sentences (\# Sent), unique sentences (\% Uniq, \# Uniq), unigrams (\# Words), and text similarity measures across various datasets. Pairwise scores (ROUGE-L, BERTScore, Tree Kernel) are computed by sampling 1,000 commands from each dataset, repeated three times for robustness. Arrows indicate increasing linguistic diversity. CR stands for Compression Ratio. The Tree Kernel method is from \citet{moschitti-2006-making}. USE refers to the minimum \# of PCA components derived from USE embeddings to explain 95\% variance for each dataset. The arrow points towards increased diversity.}
\label{tab:combined}
\end{table*}

\subsection{Instruction-Following Datasets}

To contextualize the language complexity of modern robotics datasets, we include two widely used text-only instruction-tuning datasets. To incorporate human-authored instructions, we use the English portion of the Open Assistant Conversations Dataset (OASST2) \cite{2023openassistant}. As a comparison set, we also include Alpaca \cite{alpaca}, which contains LLM-generated instructions. To represent language from visual instruction tuning, we add LLaVA-Instruct \cite{NEURIPS2023_6dcf277e}, a dataset used to train vision-language models commonly employed in VLA systems. LLaVA-Instruct primarily consists of questions about image content—such as object types, counts, actions, locations, and spatial relationships—resulting in more constrained language patterns. For all instruction-tuning datasets, we extract only the instruction texts, discarding the associated inputs and model responses.


\section{Results}
\label{sec:results}
In this section, we highlight key results of command duplication and linguistic diversity—lexical, semantic, and structural—across a range of datasets. Quantitative results allow us to identify recurring characteristics that shape the language used in embodied AI benchmarks. We focus on contrasts between VLA datasets, other robotics datasets, and instruction-tuning corpora, exploring how linguistic patterns vary across these domains. All quantitative results, specific implementation details (e.g., text preprocessing, POS-tag extraction), and additional examples are provided in the Appendices~\ref{sec:appendix-text-preprocessing}-\ref{sec:appendix-instruction-structure}.

\textbf{Analysis 1: Duplication.} Language command duplication is common across all the OXE datasets we analyzed, as well as in LIBERO. Our analysis (see Table~\ref{tab:combined}) reveals a notable disparity: in VLA datasets, fewer than 2\% of language instructions contain unique wording, which contrasts sharply with other robotics datasets and instruction-following corpora. This is largely due to the same command being paired with multiple action trajectories via multiple trials. Outside of VLA datasets, the lowest percentage of unique sentences appears in SCOUT, with around 40\% uniqueness. Many SCOUT utterances are very short—on average fewer than five words (see Figure~\ref{fig:seq_len} in the Appendix~\ref{sec:appendix-sentence-length})—which likely contributes to higher repetition compared to other non-VLA datasets. The low percentage of unique commands in OXE datasets suggests that the number of recorded episodes is not indicative of the diversity of natural language commands and highlights that during training the models frequently see the exact same commands.

\textbf{Analyses 1--3: Lexical, Semantic, and Structural Diversity Metrics.} Overall, the textual diversity metrics align: VLA datasets exhibit the lowest diversity across lexical, semantic, and structural dimensions (see Table~\ref{tab:combined}). This trend is most pronounced in the compression ratio, though pairwise scores on sampled data reveal some inconsistencies. Across pairwise metrics—ROUGE-L (lexical focus), BERTScore (semantic focus), and Tree Kernel distance (syntactic focus)—SCOUT, which includes natural human-robot dialogue, demonstrates significantly higher diversity compared to any VLA dataset. Human-written instructions from OASST2 also show considerably more diversity. For ALFRED, pairwise scores are similar to those of the more diverse VLA datasets, despite stronger results in intrinsic dimensionality and compression ratio. LLM-generated instruction-tuning datasets, particularly LLaVA-Instruct, also display limited diversity, although most metrics still report higher diversity than VLA datasets.

\textbf{Analysis 1: Lexical Diversity.} VLA datasets, particularly those with templated language, exhibit exceptionally low word counts, indicating poor lexical diversity (see Table~\ref{tab:combined}). The robotics datasets contain significantly fewer unique words than instruction-tuning datasets. For instance, RT-1, TacoPlay, and LIBERO are particularly limited in this respect—RT-1 contains only 49 unique words. In contrast, the Bridge dataset showcases the highest number of unique words among the VLA datasets, even surpassing Language Table, which includes ten times more unique commands. Non-VLA datasets like ALFRED and SCOUT also contain more unique words than most VLA datasets, despite SCOUT having fewer commands than Bridge or Language Table.

This difference with instruction-tuning datasets is somewhat expected, as instruction-following datasets are not constrained by a physical environment and can reference a broader range of objects and actions. However, some variation—such as through synonyms or paraphrasing—might be expected even in robotics contexts. Yet, many VLA datasets show little to no lexical diversity.

For example, the full set of unique words in RT-1 is limited to the following words
\begin{quote}
\textit{bottle, apple, chip, upright, green, and, pick, chocolate, open, bag, 7up, over, blueberry, shelf, rxbar, bottom, redbull, door, paper, drawer, counter, brown, knock, on, plastic, bowl, move, left, coke, fridge, blue, jalapeno, close, right, sponge, into, place, orange, pepsi, water, from, white, of, rice, top, can, near, middle, banana}.
\end{quote}

We also look into the lexical overlap between datasets. We analyze how much vocabulary is shared across datasets along the following POS categories: verbs, nouns, and adverbs (see Figure~\ref{fig:shared-tokens} in Appendix~\ref{sec:appendix-lexical-overlap}). Nouns overall are the most widely shared category, likely because many robotic tasks involve similar objects (e.g., boxes, cans, drawers). Verbs are also shared, though to a lesser extent, likely constrained by the specific capabilities of each robot embodiment. Only four words appear in all datasets: move, close, open, and pick, which are the action verbs for the majority of tasks the robots are learning to perform.

\textbf{Analysis 2: Semantic Diversity.} Compared to other robotics and instruction-tuning datasets, the difference in VLA datasets is noticeable—low intrinsic dimensionality, low word counts, and repetition point to narrow and repetitive coverage of objects, actions, and tasks. The intrinsic dimensionality analysis, measured by the number of PCA components required to explain 95\% of cumulative variance, reinforces this pattern and shows that VLA datasets are the least diverse (see Table~\ref{tab:combined}'s USE column). All four tested encoders show similar results (see Table~\ref{tab-pca-full} in Appendix \ref{sec:pca-extended}). This measure is not determined by the number of unique commands. While the number of PCA components strongly correlates with the number of unique unigrams, its relationship with the number of unique commands is notably weaker (see Figure~\ref{fig:heatmap-pca} in the Appendix~\ref{sec:pca-extended}). That further highlights the similarity between individual commands.

\begin{figure}[t!]
    \centering
    \includegraphics[width=\linewidth]{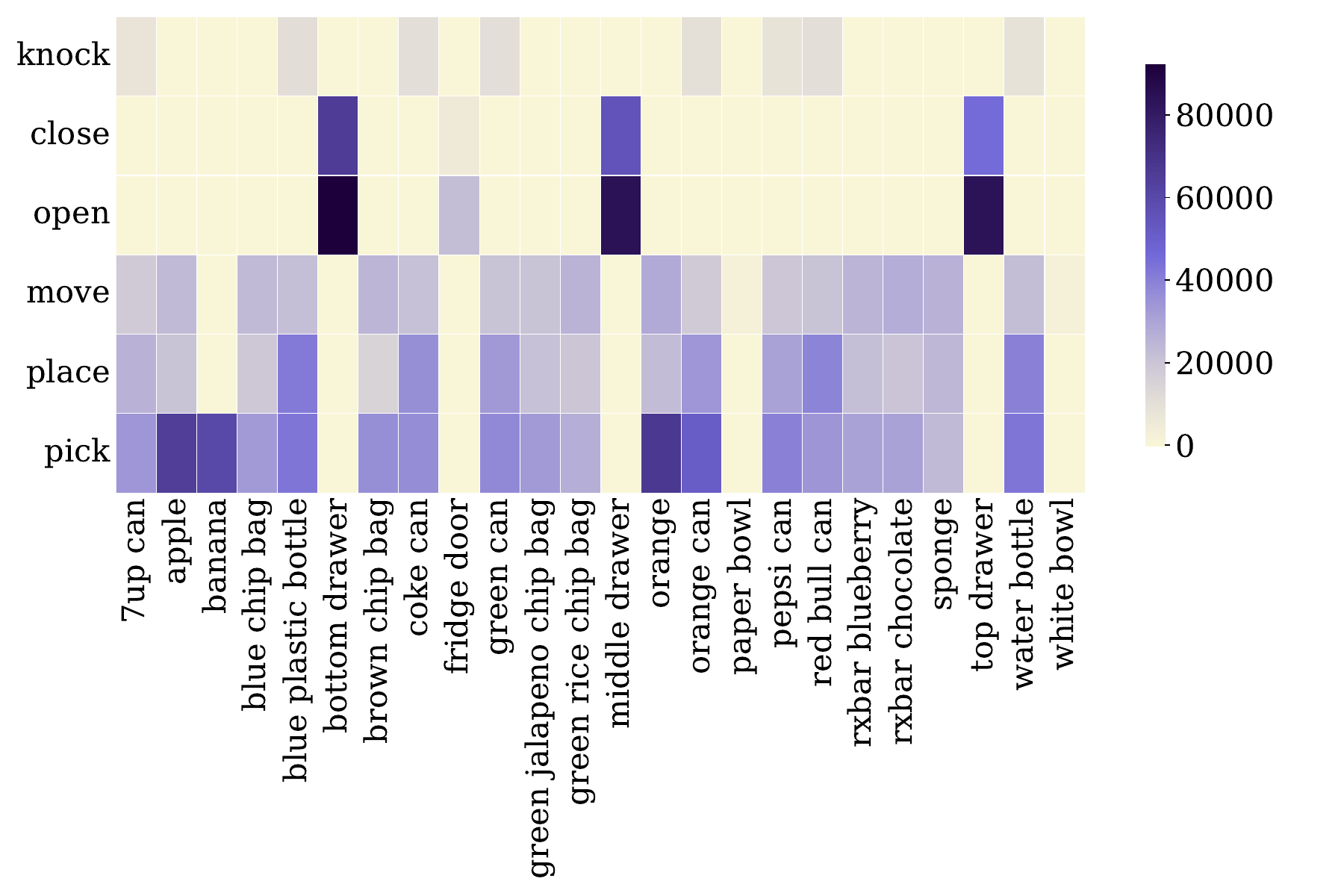}
    \caption{\textbf{Analysis 2: Semantic Diversity} Verb and direct object co-occurrence frequencies in the manually annotated RT-1 dataset. The heatmap highlights limited verb diversity across plausible actions; for instance, \textit{banana} is frequently “picked” but never “moved.” The rare verb \textit{knock} appears mostly with can-shaped objects, despite being equally applicable to others like an upright sponge.}
    \label{fig:rt1-verb-object}
\end{figure}

Even with a limited number of unique words in the RT-1 dataset, certain verb-object combinations are very frequent, such as ``pick'' and ``banana,'' whereas some objects, like ``white bowl'' and ``paper bowl'' are rarely represented at all. However, other actions involving these words occur much less frequently, which could introduce potential biases (see Figure~\ref{fig:rt1-verb-object}). Across datasets most objects co-occur with fewer than ten verbs, indicating limited task diversity (see Figures~\ref{fig:eai-dir-obj-histos} and~\ref{fig:eai-dir-obj-histos-full} in Appendix~\ref{do-method}). However, ALFRED and Language Table exhibit more balanced and varied distributions. While some constraints stem from limitations in manipulation capabilities, others appear artificial; for example, TacoPlay’s stacked blocks could support richer interactions (e.g., ``observe'' or ``tip''). For navigation datasets like SCOUT, we examine the diversity of adverbials, which modify actions in ways that convey nuance in direction, location, manner, and time. As shown in Figure \ref{fig:scout-adverbial} the lexical field is heavily skewed toward directional adverbs—notably terms such as ``north,'' ``forward,'' ``south,'' and related spatial indicators. These directionals form the backbone of navigational grounding in SCOUT; however, greater emphasis on how the robot moves, e.g., ``fast'' or ``slow'' could be beneficial.

\textbf{Analysis 3: Structural Diversity.} Multi-step instructions are frequent in robotics datasets, whereas negations, conditionals, and cyclical structures are almost entirely absent (see Figure~\ref{fig:instruct-struct} for distributions and Table~\ref{tab:instruction-examples} in the Appendix~\ref{sec:appendix-instruction-structure} for examples). Multi-step commands are the most prevalent across all datasets, reflecting a strong bias toward procedural, linear task decomposition—particularly notable in LIBERO. In contrast, datasets like RT-1 and SCOUT contain fewer multi-step instructions and favor shorter, atomic actions. Negation and conditional constructions occur in fewer than 2\% of cases, suggesting that many benchmarks fail to capture logical disjunctions, exception handling, or constraint-driven behaviors—elements essential for safe and flexible deployment. Cyclical or loop-like structures, which are common in real-world tasks, are similarly underrepresented, with only SCOUT and ALFRED showing a modest signal. Overall, these patterns point to a structural bias in current datasets toward flat, step-by-step formulations, with limited support for complex task logic.

\newcommand{\verbtag}{\textcolor{ALFRED}{\textbf{VERB}}}
\newcommand{\nountag}{\textcolor{LIBERO}{\textbf{NOUN}}}
\newcommand{\adjtag}{\textcolor{RTOne}{\textbf{ADJ}}}
\newcommand{\advtag}{\textcolor{SCOUT}{\textbf{ADV}}}
\newcommand{\adptag}{\textcolor{TacoPlay}{\textbf{ADP}}}

\begin{figure}[t!]
\scriptsize
\noindent
\textbf{RT-1 }\cite{brohan2023rt1roboticstransformerrealworld} \\[0.9ex]
\scriptsize\quad \verbtag $\rightarrow$ \nountag $\rightarrow$ \nountag $\rightarrow$ \adptag $\rightarrow$ \adjtag $\rightarrow$ \nountag \hfill (11\%)\\
place water bottle into white bowl \\[0.5ex]
\scriptsize\quad \verbtag $\rightarrow$ \nountag $\rightarrow$ \nountag $\rightarrow$ \adptag $\rightarrow$ \nountag $\rightarrow$ \nountag \hfill (7\%)\\
move redbull can near 7up can \\[0.5ex]
\vspace{0.4ex}\rule{\linewidth}{0.1pt}\vspace{0.9ex}
\scriptsize \textbf{BRIDGE} \cite{bridge_2022} \\[0.9ex]
\scriptsize\quad \verbtag $\rightarrow$ DET $\rightarrow$ \nountag $\rightarrow$ \adptag $\rightarrow$ DET $\rightarrow$ \nountag $\rightarrow$ PUNCT \hfill (3\%)\\
Place the mushroom behind the spatula. \\[0.5ex]
\scriptsize\quad \verbtag $\rightarrow$ DET $\rightarrow$ \nountag $\rightarrow$ \adptag $\rightarrow$  DET $\rightarrow$ \nountag $\rightarrow$ \adptag $\rightarrow$ \\ DET $\rightarrow$ \nountag $\rightarrow$ PUNCT \hfill (3\%)\\
Move the spoon to the left of the napkin. \\[0.5ex]
\vspace{0.4ex}\rule{\linewidth}{0.1pt}\vspace{0.9ex}
\scriptsize \textbf{TacoPlay} \cite{rosete2022tacorl} \\[0.9ex]
\scriptsize\quad \verbtag $\rightarrow$ DET $\rightarrow$ \adjtag $\rightarrow$ \nountag $\rightarrow$ \adptag $\rightarrow$ DET $\rightarrow$ \nountag \hfill (24\%)\\[0.5ex]
put the purple block on the table \\[0.5ex]
\scriptsize\quad \verbtag $\rightarrow$ DET $\rightarrow$ \adjtag $\rightarrow$ \nountag $\rightarrow$ \adptag $\rightarrow$ DET $\rightarrow$ \adjtag $\rightarrow$ \nountag \hfill (6\%)\\[0.5ex]
put the pink object inside the left cabinet  \\[0.5ex]
\vspace{0.4ex}\rule{\linewidth}{0.1pt}\vspace{0.9ex}
\scriptsize \textbf{Language Table} \cite{langtable} \\[0.9ex]
\scriptsize\quad \verbtag $\rightarrow$ DET $\rightarrow$ \adjtag $\rightarrow$ \nountag $\rightarrow$ \advtag $\rightarrow$ \adptag $\rightarrow$ DET $\rightarrow$ \adjtag $\rightarrow$ \nountag \hfill (4\%)\\[0.5ex]
move the blue cube right to the yellow hexagon \\[0.5ex]
\scriptsize\quad \verbtag $\rightarrow$ DET $\rightarrow$ \adjtag $\rightarrow$ \nountag $\rightarrow$ \adptag $\rightarrow$ DET $\rightarrow$ \adjtag $\rightarrow$ \nountag \hfill (4\%)\\[0.5ex]
push the green circle towards the green star \\[0.5ex]
\caption{\textbf{Analysis 3: Structural Diversity} Most frequent POS patterns per dataset on unique sentences and their relative frequency and example.}
\label{fig:pos-patterns}
\end{figure}
\pgfplotstableread{
0 0.02   0.00
1 0.22   0.00
2 44.33   0.85
3 2.62   0.31
}\mergedALFRED

\pgfplotstableread{
0 1.36   0.00
1 0.95   0.00
2 21.07   0.74
3 4.23   0.21
}\mergedSCOUT

\pgfplotstableread{
0 0.00    0.00
1 0.00    0.00
2 13.71    0.00
3 0.00    0.00
}\mergedRTOne

\pgfplotstableread{
0 0.23    0.00
1 0.02    0.00
2 26.33    1.05
3 1.18   0.26
}\mergedBRIDGE

\pgfplotstableread{
0 0.00    0.00
1 0.00    0.00
2 25.81    0.00
3 0.00    0.00
}\mergedTacoPlay

\pgfplotstableread{
0 0.02    0.00
1 0.00    0.00
2 17.37    0.74
3 3.59    0.33
}\mergedLanguageTable

\pgfplotstableread{
0 0    0.00
1 0    0.00
2 56.66    0.00
3 0    0.00
}\mergedLIBERO

\begin{figure}[ht!]
\centering
\begin{tikzpicture}
\begin{axis}[
    width=0.5\textwidth,
    height=4.5cm,
    major x tick style = transparent,
    x axis line style = {opacity=0},
    y axis line style = {opacity=0},
    axis y line = left,
    ylabel near ticks,
    ymajorgrids = true,
    ytick style={draw=none},
    ylabel = {\% of Instructions},
    ylabel style={font=\scriptsize},
    xtick = {0, 1, 2, 3},
    xticklabels = {Negation, Conditional, Multi Step, Cycle},
    xticklabel style = {align=center, font=\scriptsize},
    yticklabel style = {font=\scriptsize},
    ymin=0, ymax=70,
    enlarge x limits=0.1,
    /pgf/bar width={5pt},  
    ybar=1.25pt,
    legend cell align=left,
    legend style={
        at={(0.425,1.00)},
        anchor=south,
        font=\scriptsize,
        draw=none,
        legend columns=4,
        column sep=0.5ex,
        inner xsep=2pt,
        inner ysep=2pt,
        nodes={scale=0.8, transform shape}
    }
]

\addplot+[
    style={ALFRED, fill=ALFRED, mark=none},
    draw=none, fill opacity=0.9,
    error bars/.cd, y dir=both, y explicit
] table[x index=0, y index=1, y error index=2] \mergedALFRED;

\addplot+[
    style={SCOUT, fill=SCOUT, mark=none},
    draw=none, fill opacity=0.9,
    error bars/.cd, y dir=both, y explicit
] table[x index=0, y index=1, y error index=2] \mergedSCOUT;

\addplot+[
    style={RTOne, fill=RTOne, mark=none},
    draw=none, fill opacity=0.9,
    error bars/.cd, y dir=both, y explicit
] table[x index=0, y index=1, y error index=2] \mergedRTOne;

\addplot+[
    style={BRIDGE, fill=BRIDGE, mark=none},
    draw=none, fill opacity=0.9,
    error bars/.cd, y dir=both, y explicit
] table[x index=0, y index=1, y error index=2] \mergedBRIDGE;

\addplot+[
    style={TacoPlay, fill=TacoPlay, mark=none},
    draw=none, fill opacity=0.9,
    error bars/.cd, y dir=both, y explicit
] table[x index=0, y index=1, y error index=2] \mergedTacoPlay;

\addplot+[
    style={LanguageTable, fill=LanguageTable, mark=none},
    draw=none, fill opacity=0.9,
    error bars/.cd, y dir=both, y explicit
] table[x index=0, y index=1, y error index=2] \mergedLanguageTable;

\addplot+[
    style={LIBERO, fill=LIBERO, mark=none},
    draw=none, fill opacity=0.9,
    error bars/.cd, y dir=both, y explicit
] table[x index=0, y index=1, y error index=2] \mergedLIBERO;

\legend{ALFRED, SCOUT, RT-1, BRIDGE, TacoPlay, LanguageTable, LIBERO}

\end{axis}
\end{tikzpicture}
\caption{\textbf{Analysis 3: Structural Diversity} Percentage of instructions exhibiting four structural phenomena: negation, conditionality, multi-step sequencing, and cyclic repetition. For datasets containing fewer than 600 unique sentences, annotations were performed manually. For those with more than 600 unique sentences, annotations were generated using an automated pipeline. Standard error bars reflect labeling uncertainty estimated from a manually reviewed subset of 500 randomly sampled commands per dataset.}
\label{fig:instruct-struct}
\end{figure}
Analyzing part-of-speech (POS) patterns yields similar insights. This analysis examines the grammatical structure of commands, specifically focusing on how words are arranged using POS patterns. The most frequent patterns typically start with a verb and describe the object using either a single noun, an adjective followed by a noun, or two nouns, followed by an adposition and a new object description. In the cases of RT-1 and TacoPlay, the most frequent pattern constitutes 11\% and 24\% of all patterns, respectively (see Figure~\ref{fig:pos-patterns}). This reliance on repetitive structures may make it harder for models to generalize to more complex instructions. Refer to Figures~\ref{tab:pos-patterns-examples1} and~\ref{tab:pos-patterns-examples2} in Appendix~\ref{pos-method} for more qualitative examples of dominant patterns. Figure~\ref{fig:dep-parse-histo} offers an aggregated view across datasets, and Figures~\ref{fig:by-ds-dep-histo-grid-1} and \ref{fig:by-ds-dep-histo-grid-2} group by dataset.

The command length distribution across seven datasets reveals a preference for short commands that fall within the range of 3 to 15 words. This further highlights the dominance of concise phrasing, which may limit exposure to more complex linguistic structures, e.g., multi-clause, multi-step instructions.

\section{Conclusion \& Future Directions}
\label{sec:conclusion}
In this work, we characterize linguistic diversity in widely used VLA datasets using complementary lexical, semantic, and structural metrics. Across VLA datasets, the language is highly repetitive: there are few unique commands, limited lexical variation, and narrow coverage of objects, actions, and tasks. Syntactic diversity is also constrained: multi-step commands are common, while negation and conditionals are rare. 

Prior work suggests that linguistic diversity may matter for generalization. Our analyses provide a starting point for doing so in VLA datasets by identifying which aspects of language are currently underrepresented. We can leverage these insights to improve linguistic diversity using several possible strategies: (i) \textbf{Targeted augmentation}—using paraphrasing and template-based generation to expand underrepresented patterns; (ii) \textbf{Cross-domain transfer}—selectively leveraging linguistically richer instructional corpora (e.g., procedural text, situated dialogue) to complement VLA data; and (iii) \textbf{Annotation guidance}—encouraging data collection that explicitly includes missing constructions. Among these, our analyses most directly support both (i) targeted augmentation and (iii) annotation practices. 

Targeted augmentation enables us to improve existing datasets' linguistic diversity without building new datasets. Based on our analyses, we can leverage syntactic similarity metrics—such as TreeKernel methods or POS pattern histograms (Analysis 3)—to guide LLMs in generating paraphrases through phrase permutation or inversion. In many cases, the vocabulary within current datasets is limited (Analysis 1). Here, LLM-guided synonym replacement could help increase lexical diversity. 

A more resource-intensive—but higher-quality—approach is to use these findings to guide future data collection. For example, Analysis 1 reveals strong n-gram overlap and frequent word duplication. To mitigate this, data collection interfaces could prompt users in real time to rephrase instructions using alternative wording or synonyms. Analysis 2 shows strong co-occurrence patterns between specific verbs and nouns in the RT-1 dataset, suggesting the presence of potentially superficial correlations. Addressing this may require encouraging more varied verb–object pairings during data collection. Finally, Analysis 3 highlights that POS patterns are highly clustered within datasets and that structural diversity is dominated by multi-step commands. To counteract this, participants could be prompted to use more descriptive and varied language—for example, by incorporating adverbial phrases—since the repetitive nature of dataset construction may otherwise discourage linguistic richness. Our analysis also suggests that more interactive data collection methods—such as SCOUT’s Wizard-of-Oz approach—can yield greater lexical diversity (Analysis 1) and richer structural variation (Analysis 3). This points to a promising pathway for generating linguistically diverse data.

\section{Limitations}
\label{sec:limitations}
Our analysis is motivated by the goal of enabling more generalist policies and stronger robotics foundation models, with language as a core modality. The patterns we identify may be less critical in narrower settings or application domains with constrained task distributions. Moreover, not all linguistic phenomena are equally relevant in all pipelines: for example, negation may be crucial for more atomic commands, whereas conditionals may matter less when planning is delegated to an external LLM.

We also acknowledge that object diversity is often limited by the costs of acquiring props and training novel tasks which biases robotics datasets toward low linguistic diversity. However, our work points toward the possibility that certain object categories are overrepresented--indicating that investment in new prop objects or alternative environment scenery (e.g., beyond kitchen environments) may be more beneficial.

Although robotics datasets are inherently multimodal, our study focuses exclusively on their textual components. Language functions in tandem with visual input and action trajectories in embodied systems; because we do not evaluate cross-modal alignment, our findings do not capture potential inconsistencies between commands and the corresponding trajectories or visual scenes.

Individual metrics also have known limitations and may not capture linguistic diversity in full. To mitigate this, we report a diverse set of complementary lexical, semantic, and structural measures, and supplement them with manual analyses and multiple methodological perspectives where appropriate.

Finally, while our study covers several widely used VLA and robotics datasets that reflect dominant trends in the field, the results may not generalize to all instruction-following datasets in embodied AI.

\section*{Acknowledgments}

This study was in part funded by the Estonian Research Council grants PRG3237 and PRG2006; and LDRD grant 20250048DR (U.S. DOE NNSA
Contract 89233218CNA000001) (LA-UR-25-24853).

\bibliography{custom}

\appendix

\section{Qualitative Features of EAI datasets}
\label{sec:qual-feat}

We conducted an informal qualitative review of the examined datasets and highlighted interesting attributes, summarized in Table \ref{tab:qual_examples}.

\begin{table*}[h!]
\centering
\small
\begin{tabular}{p{5.5cm} p{9cm}}
\toprule
\textbf{Theme} & \textbf{Example Instruction(s)} \\
\midrule
Cultural Terms (BRIDGE) & ``put the kadai on the stove'', ``grab the brinjal from the drawer'' \\
Unsafe Action (ALFRED) & ``store a knife in a microwave'', ``stab the tip of the knife into the table'' \\
Commonsense Violation (ALFRED) & ``Put an egg in a pan in the fridge'' \\
Commonsense Violation (BRIDGE) & ``take sushi out of the pan'' \\
\bottomrule
\end{tabular}
\caption{Selected examples illustrating conversational structure, cultural variation, and commonsense inconsistencies across EAI datasets.}
\label{tab:qual_examples}
\end{table*}

\textbf{On Conversational Strengths.}
The SCOUT dataset exhibits a distinct dialogue structure that differentiates it from traditional instruction-following datasets. Rather than adhering to a rigid, directive style, its dialogues often involve an exploratory or inquiry-based approach, as seen in exchanges like ``move west uh zero point five meters'' and ``...and then the last question here anything that indicates the environment was recently occupied''. This interactive nature may offer advantages for EAI by allowing more adaptive responses. For example, in cases where instructions involve complex spatial reasoning; e.g., placing an object in a specific but ambiguous location, the dataset’s conversational format could aid in disambiguation.

\textbf{On Cultural Knowledge.}
One of the more striking aspects of the BRIDGE dataset is its incorporation of multicultural culinary terminology, despite being primarily monolingual (English). Unlike many Western-centric datasets, BRIDGE includes references to diverse cooking utensils and ingredients, such as purkoli (broccoli), brinjal (eggplant), brezzela (eggplant), capsicum (bell pepper), quince fruit, nigiri, wok, and kadai. This linguistic diversity suggests a broader representation of cultural knowledge, making incremental progress toward addressing concerns raised in prior work on dataset biases \cite{stochastic-parrots, bender2019benderrule}. Specifically, it challenges the tendency for data collection to reflect primarily Western, white audiences. Additionally, BRIDGE captures subtle social characteristics of human perception, such as humor, evidenced by an annotation that describes a mushroom toy as a “phallic looking item.''

\textbf{On ``Common Sense'' Reasoning.}
A recurring challenge across real-world datasets is the disconnect between world knowledge, common-sense reasoning, and practical instruction execution. While BRIDGE and ALFRED aim to ground tasks in realistic environments, many instructions contain fundamental inconsistencies or implausible directives. In ALFRED, for example, commands such as ``open refrigerator, place potato to the right of tomato on second shelf of refrigerator, close refrigerator, open refrigerator, pick up potato from refrigerator, close refrigerator'' expose rigid, mechanical assumptions about human behavior. Additionally, one must ask what has been accomplished by storing a potato in a refrigerator and then removing said potato in a matter of seconds. Another example from ALFRED includes, ``Put an egg in a pan in the fridge.'' More concerning, and at times, unintentionally amusing, are instances of potentially unsafe or property-damaging instructions, such as ``place a heated slice of tomato on a counter and \textbf{store a knife in a microwave}'' or ``\textbf{stab the tip of the knife into the wooden table}, in front of the gray plate closest to the lettuce.'' While a robot damaging a kitchen table may be preferable to microwaving a knife, these examples highlight inconsistencies in world knowledge modeling within these datasets. Similar anomalies appear in BRIDGE, where commands such as ``take sushi out of the pan,'' ``put sushi in pot...,'' and ``put spatula in pan'' suggest an oversimplified understanding of object affordances, human behavior, and broader world and cultural knowledge. If the broader EAI community sees embodiment as a necessary step toward elevating the representational learning of single-modality models, e.g., LLMs, we ought to discourage dataset collectors from building illogical ``common-sens''" associations.

\section{Text Preprocessing}
\label{sec:appendix-text-preprocessing}

For calculating the metrics in Table~\ref{tab:combined}, we preprocess the text by splitting the text into sentences, standardizing white space, and removing punctuation. We calculate the statistics using the combination of \texttt{spacy} \cite{spacy} and \texttt{pandas} \cite {reback2020pandas} methods. 

To measure unique words, we removed punctuation from our cleaned text. Then, we concatenated all sentences in a particular corpus into a single string. We tokenized the text into unigrams using Python’s \texttt{.split()}, converted the resulting list into a \texttt{.set()} to obtain unique tokens, and computed the number of unique words as the size of that set.

For all experiments, we cleaned the SCOUT~\cite{lukin-etal-2024-scout-situated} dataset of user role tags and tags that indicate filler words, e.g., ``um'', silence, and noise. Due to the complexity of this data, we focus our initial analysis only on the ``robot commander'' dialogue, with plans to expand our analysis to all roles in the future and to incorporate filler filtering in the text cleaning pipeline. 

\section{Lexical Diversity Extended Results}
\label{sec:appendix-lexical-diversity}

In addition to the metrics discussed in the main body, we measured sentence length (Section~\ref{sec:appendix-sentence-length}), lexical overlap  (Section~\ref{sec:appendix-lexical-overlap}), and some additional lexical diversity metrics (Section~\ref{sec:appendix-additional-metrics}). 

\subsection{Sentence Length}
\label{sec:appendix-sentence-length}

\begin{figure}[h!]
    \centering
    \includegraphics[width=\linewidth]{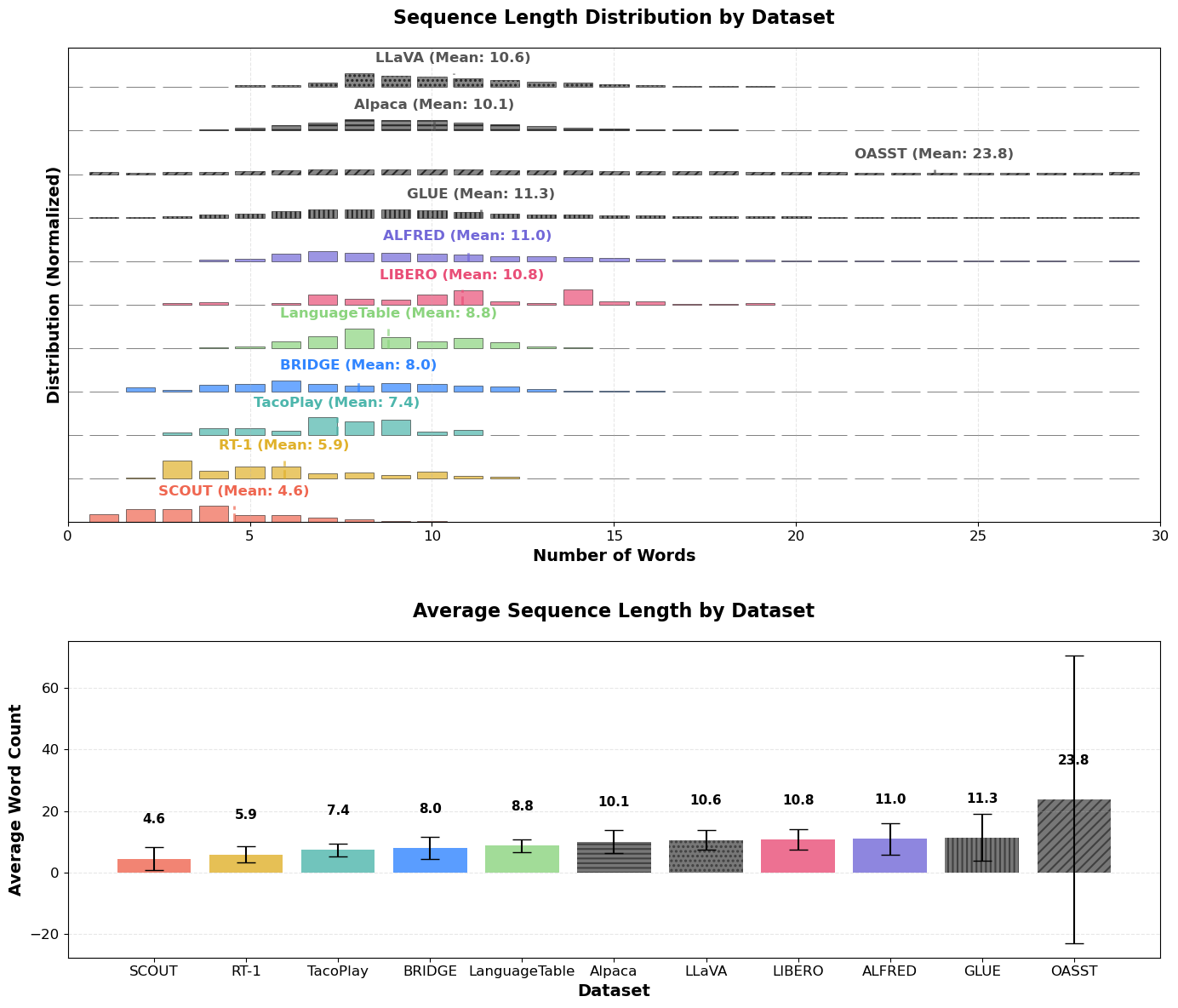}
    \caption{\textbf{Analysis 1: Lexical Diversity} Distribution of command lengths across six examined EAI datasets.}
    \label{fig:seq_len}
\end{figure}

The majority of commands contain fewer than ten words (see Figure \ref{fig:seq_len}). Command lengths are capped at a maximum of 30 words for display purposes. OASST shows wide standard deviation because of outlier examples from the instruction datasets that contain up to 100+ words. Despite the impossibility of a negative sentence length we display the standard deviation exactly as other datasets. The \textit{Instruction-Tuning and NLU Datasets} are all shown in grayscale with various hatching to distinguish. Generally, these datasets share similar sentence lengths as Language Table, LIBERO, ALFRED, however greatly outperform the aforementioned robotics datasets against our diversity measures (c.f. Tables \ref{tab:combined} and \ref{tab-pca-full}).

\subsection{Lexical Overlap}
\label{sec:appendix-lexical-overlap}

\begin{figure}[h]
    \centering
    \includegraphics[width=\linewidth]{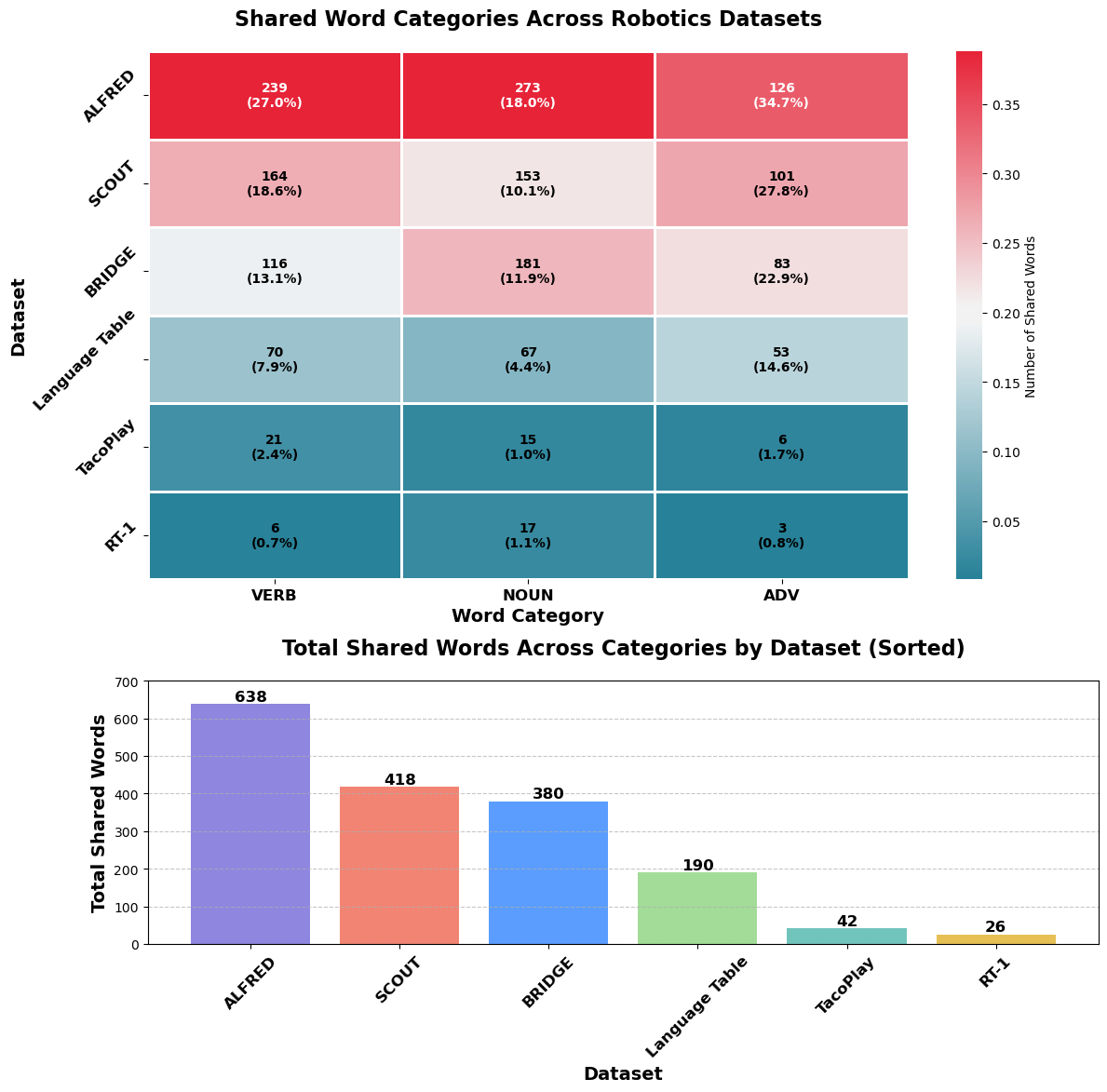}
    \caption{\textbf{Analysis 1: Lexical Diversity} Shared POS categories across datasets. Using ALFRED as a pretraining dataset is advantageous because it has the greatest amount of lexical coverage across the examined EAI datasets.}
    \label{fig:shared-tokens}
\end{figure}

To assess how much vocabulary is shared across datasets, we examine the distribution of words across three part-of-speech (POS) categories: nouns, verbs, and adverbs. We use dependency parsing to extract tokens by their POS tags.  We then construct a dataset–word matrix that records how often each word appears in more than one dataset. This allows us to visualize lexical overlap using a heatmap (Figure~\ref{fig:shared-tokens}). 

\subsection{Additional Metrics: Levenshtein, Jaccard, BLEU-4}
\label{sec:appendix-additional-metrics}

In addition to metrics shown in~\ref{tab:combined}, we measured  Levenshtein, Jaccard, and, following previous work~\cite{ijcai2020p0124}, BLEU-4.  Given that these methods entail pair-wise comparisons, we perform 1,000 commands to obtain these scores across 3 trials, as with other pair-wise metrics. The results are in Table~\ref{tab-text-sim-ext}, and the scores agree with other metrics.

\begin{table*}
\begin{center}
\small
\setlength{\tabcolsep}{1mm}

\begin{tabular}{lccc}
\toprule
Dataset & Levenshtein $\uparrow$ & Jaccard $\downarrow$ & BLEU-4 $\downarrow$ \\
\midrule
\multicolumn{4}{l}{\textit{Instruction-Tuning Datasets}} \\
\hspace{3mm}OASST2 \cite{2023openassistant} & \textbf{78.238 $\pm$ 1.789} & \textbf{0.033  $\pm$ 0.001} & \textbf{0.0001 $\pm$ 0.0001} \\ 
\hspace{3mm}Alpaca \cite{alpaca} & 51.382 $\pm$ 0.537 & 0.061 $\pm$ 0.001 & 0.0003 $\pm$ 0.0001 \\
\hspace{3mm}LLaVA-Instruct \cite{NEURIPS2023_6dcf277e} & 46.465 $\pm$ 0.223 & 0.130 $\pm$ 0.002 & 0.002 $\pm$ 0.001 \\
\midrule
\multicolumn{4}{l}{\textit{Language-Focused Robotics Datasets}} \\
\hspace{3mm}ALFRED \cite{Shridhar_2020_CVPR} & \textbf{46.695 $\pm$ 0.883} & 0.128 $\pm$ 0.004 & 0.003 $\pm$ 0.000 \\
\hspace{3mm}SCOUT \cite{lukin-etal-2024-scout-situated} & 24.512 $\pm$ 0.946 & \textbf{0.052 $\pm$ 0.002} & \textbf{0.002 $\pm$ 0.001} \\
\midrule
\multicolumn{4}{l}{\textit{VLA Datasets}} \\
\hspace{3mm}RT-1 \cite{brohan2023rt1roboticstransformerrealworld} & 28.143 $\pm$ 0.413 & 0.138 $\pm$ 0.001 & 0.026 $\pm$ 0.006 \\
\hspace{3mm}BRIDGE \cite{walke2023bridgedata} & \textbf{35.139 $\pm$ 0.180} & \textbf{0.088 $\pm$ 0.004} & \textbf{0.003 $\pm$ 0.000} \\
\hspace{3mm}TacoPlay \cite{rosete2022tacorl} & 27.705 $\pm$ 0.137 & 0.188 $\pm$ 0.003 & 0.020 $\pm$ 0.001 \\
\hspace{3mm}Language Table \cite{langtable} & 32.206 $\pm$ 0.171 & 0.198 $\pm$ 0.002 & 0.010 $\pm$ 0.001 \\
\hspace{3mm}LIBERO \cite{liu2023liberobenchmarkingknowledgetransfer} & 34.269 $\pm$ 0.188 & 0.248 $\pm$ 0.006 & 0.064 $\pm$ 0.003 \\
\bottomrule
\end{tabular}
\end{center}
\caption{\textbf{Analysis 1: Lexical Diversity} Subset of text similarity measures: Levenshtein distance, Jaccard similarity, and BLEU-4. Arrows indicate that higher Levenshtein and lower Jaccard/BLEU-4 correspond to greater diversity.}
\label{tab-text-sim-ext}
\end{table*}

\section{Intrinsic Dimensionality Analysis}
\label{sec:pca-extended}

A notable limitation of our methodology is using linear dimensionality reduction techniques, specifically PCA, to assess data that may lie on a nonlinear manifold, as is often the case with LLM-encoded datasets. While PCA assumes linearity, this limitation does not significantly undermine our analysis. In fact, it likely results in an \textit{overestimation} of the intrinsic dimensionality, since PCA cannot exploit underlying nonlinear relationships in the data \cite{intrinsic-dim-non-lin-pca}. For our purposes, this effect only further underscores the discrepancy between the structure of robotics datasets and the more diverse language representations found in other datasets we studied.

\begin{figure}[h]
    \centering
    \includegraphics[width=\linewidth]{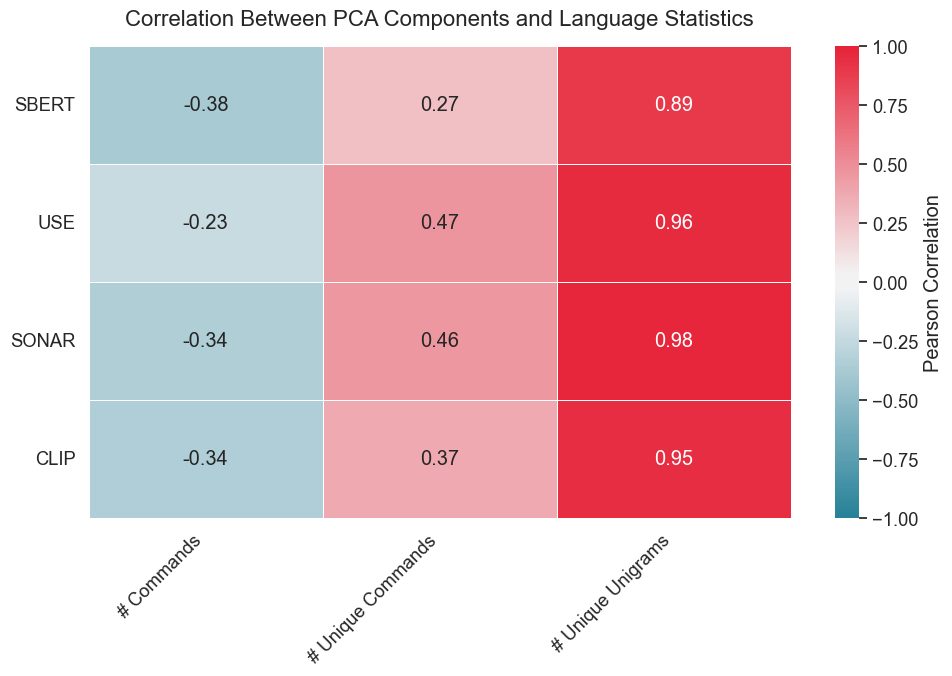}
    \caption{Correlation between the number of PCA components required to explain 95\% variance and language statistics across EAI datasets.
PCA components derived from SBERT, USE, SONAR, and CLIP embeddings are compared against the number of commands, unique commands, and unique unigrams in each dataset. Strong positive correlations are observed between unique unigrams and all embedding models, particularly SONAR and USE. In contrast, the total number of commands shows weak or negative correlation with embedding diversity}
    \label{fig:heatmap-pca}
\end{figure}

The full results with four different sentence encoders are in Table~\ref{tab-pca-full}. We also measured the correlation between the number of PCA components required to explain 95\% variance and language statistics across EAI datasets (see Figure~\ref{fig:heatmap-pca}).

Although the conclusions of this analysis are reinforced by our more interpretable feature-based methods (see Table \ref{tab:combined}); in future work, we would like to strengthen this effort. 

\begin{table}[t!]
\begin{center}
\scriptsize
\setlength{\tabcolsep}{0.5mm}
\begin{tabular}{lcccc}
\toprule
Dataset & SBERT $\uparrow$ & USE $\uparrow$ & SONAR $\uparrow$ & CLIP $\uparrow$\\
\midrule
\multicolumn{5}{l}{\textit{Instruction-Tuning}} \\
\hspace{1mm}OASST2 \cite{2023openassistant} & \textbf{396} & 254 & \textbf{754} & 361 \\ 
\hspace{1mm}Alpaca \cite{alpaca} & 350 & 231 & 637 & 338 \\
\hspace{1mm}LLaVA-Instruct \cite{NEURIPS2023_6dcf277e} & 245 & 184 & 540 & 279 \\
\midrule
\multicolumn{5}{l}{\textit{Language-Focused Robotics Datasets}} \\
\hspace{1mm}ALFRED \cite{Shridhar_2020_CVPR} & 165 & \textbf{159} & \textbf{406} & \textbf{198} \\
\hspace{1mm}SCOUT \cite{lukin-etal-2024-scout-situated} & \textbf{194} & 148 & 295 & 181 \\
\midrule
\multicolumn{5}{l}{\textit{VLA Datasets}} \\
\hspace{1mm}RT-1 \cite{brohan2023rt1roboticstransformerrealworld} & 27 & 33 & 42 & 35\\
\hspace{1mm}BRIDGE \cite{walke2023bridgedata} & \textbf{115} & \textbf{125} & \textbf{239} & \textbf{149} \\
\hspace{1mm}TacoPlay \cite{rosete2022tacorl} & 31 & 42 & 41 & 36 \\
\hspace{1mm}Language Table \cite{langtable} & 57 & 86 & 108 & 71 \\
\hspace{1mm}LIBERO \cite{liu2023liberobenchmarkingknowledgetransfer} & 32 & 34 & 44 & 33 \\
\bottomrule
\end{tabular}
\end{center}
\caption{\textbf{Analysis 2: Semantic Diversity} The minimum number of PCA components to explain 95\% variance for each dataset. A greater number of components represents stronger diversity.}
\label{tab-pca-full}
\end{table}

\section{POS Patterns}
\label{pos-method}

\begin{figure}
    \centering
    \includegraphics[width=\linewidth]{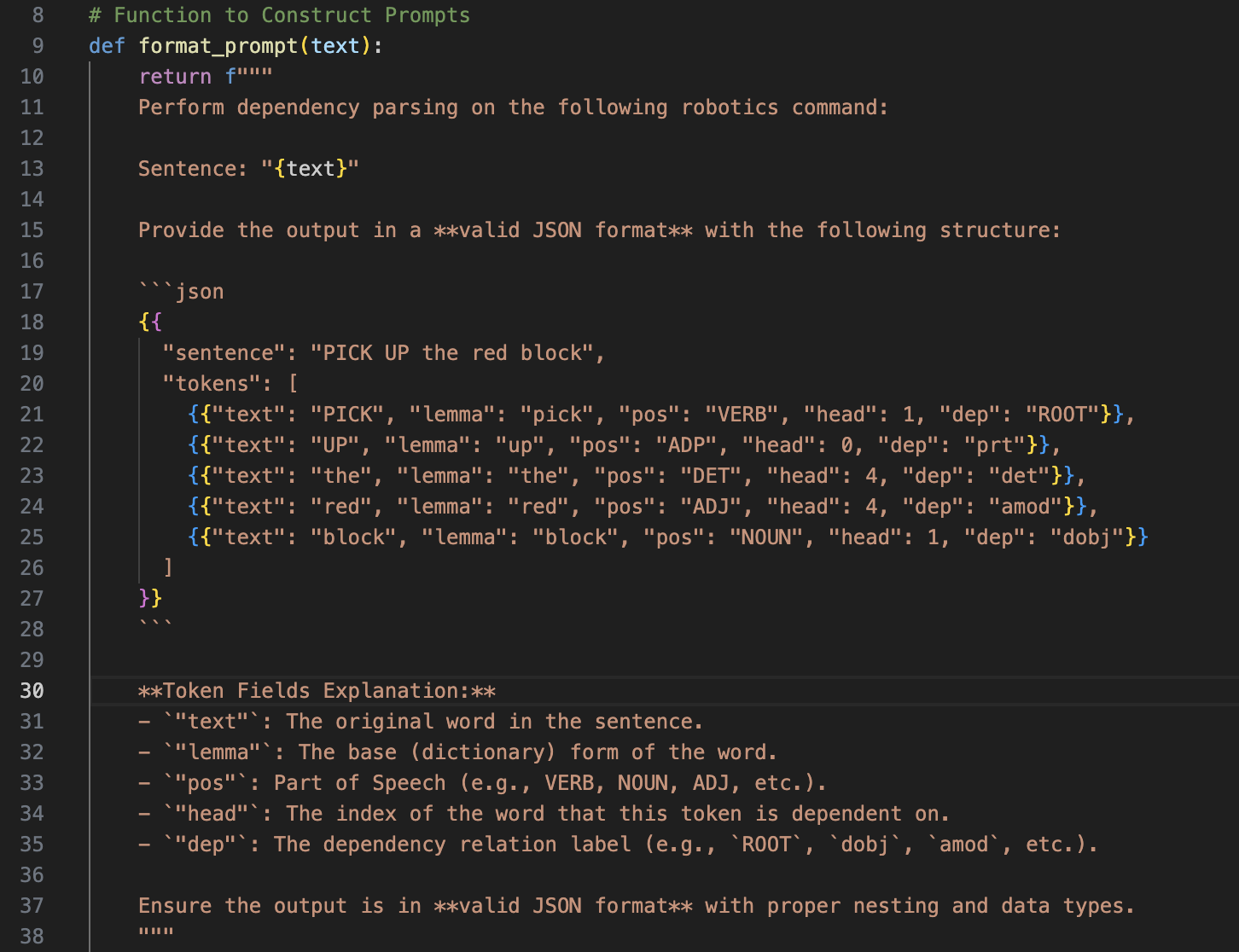}
    \caption{Prompt used in dependency parse work.}
    \label{fig:dep-parse-prompt}
\end{figure}

We implemented a large-scale dependency parsing pipeline using an LLM to extract POS and dependency parse patterns, leveraging multi-GPU parallel processing for efficiency. Each GPU independently processed a subset of instructions using \texttt{DeepSeek-R1-Distill-Qwen-32B} \cite{deepseekai2025deepseekr1}, a state-of-the-art instruction-following LLM. The model was loaded in 8-bit quantized format to optimize memory usage, and batch \(b = 10\) processing was employed to maximize throughput. The prompts for the model followed a structured format (see Figure \ref{fig:dep-parse-prompt}), instructing it to perform dependency parsing and return results in valid JSON format. The output JSON included:

\begin{itemize}
    \item The original instruction
    \item A tokenized breakdown, where each word was annotated with its:
    \begin{itemize}
        \item Lemma (root form)
        \item Part of speech (POS) tag
        \item Syntactic head (parent word in the dependency tree)
        \item Dependency label (e.g., ROOT, direct object, modifier, etc.)
    \end{itemize}
\end{itemize}

For qualitative examples related to each POS pattern, please refer to Figures \ref{tab:pos-patterns-examples1} and \ref{tab:pos-patterns-examples2}.

The decision to use LLMs for the POS tagging task was driven by an exploratory, qualitative review of preliminary outputs from several traditional NLP tools. We evaluated spaCy models, including \texttt{en\_core\_web\_sm} and \texttt{en\_core\_web\_trf}, as well as StanfordNLP’s Stanza POS tagging model (\texttt{POSProcessor}). Across these models, including transformer-based variants, we observed recurring difficulties with the unconventional object naming conventions in the RT-1 dataset (e.g., \texttt{rxbar}). Several models also struggled to reliably distinguish between the noun ``can'' (e.g., referring to 7up cans in the dataset) and the verb ``can'' (expressing ability). During informal testing, the DeepSeek models appeared to handle these cases better more consistently; so we opted to build our POS tagging workflow around them. However, to emphasize, we did not conduct a quantitative comparison across parsers.

The \textbf{BRIDGE} dataset is heavily characterized by prepositional phrases, frequently structuring instructions that specify spatial relationships between objects and the environment. This results in a high frequency of ADP (adpositions), NOUN (nouns), and DET (determiners), forming patterns, e.g. ``put the spoon on the cloth'', ``put the mangoes in a pan'', and ``Move the spatula near the egg.'' While this structure ensures precision in command execution, it lacks syntactic variation beyond simple prepositional constructs, potentially limiting generalization to more complex spatial reasoning tasks. 

\textbf{RT-1}, in particular, exhibits highly repetitive syntactic patterns, as seen in commands like ``place 7up can into middle drawer,'' ``place water bottle into white bowl,'' and ``place rxbar blueberry into bottom drawer.'' Similarly, TacoPlay demonstrates significant syntactic redundancy, with instructions such as ``place the purple block on the table,'' ``store the pink object in the drawer,'' and ``slide the yellow block to the right.'' This lack of linguistic variability, likely due to the template-driven generation of these datasets, may limit a model’s ability to generalize to more complex instructions, particularly those involving hierarchical dependencies or compound actions. 

\textbf{SCOUT} introduces more numerical expressions and adverbial structures, implying an instructional style where robots may be required to count, measure, or modify behaviors dynamically, e.g., ``move south four feet'', ``turn right twenty degrees'', ``go forward one meter''. However, its emphasis on concise command structures might underrepresent more complex multi-step directives.  

The POS histograms in Figures \ref{fig:dep-parse-histo}-\ref{fig:by-ds-dep-histo-grid-2} reveal a long-tailed distribution in TacoPlay, SCOUT, and RT-1, where the frequency of syntactic structures drops sharply after the first or second most common parse pattern. Such patterns indicate a reliance on repetitive syntactic templates, which may limit a model’s ability to generalize to linguistically varied instructions. Language Table shows the longest and most evenly distributed bar set among all datasets, with no single POS pattern dominating. Language Table sets the upper bound for linguistic diversity among embodied AI datasets and should be more widely used. However, for datasets like RT-1, we recommend that synthetic data augmentation could help mitigate this imbalance by introducing greater syntactic variability, such as tree-based reordering techniques, inspired by data augmentation in machine translation \cite{dehouck-gomez-rodriguez-2020-data,shi2021substructuresubstitutionstructureddata}, could be adapted to generate syntactic variants of robotic commands while preserving their semantics.

\textbf{On Annotation Quality.} Table \ref{tab:pos_standard_error_rates} presents standard error rates on POS tagging using \texttt{GPT-5.2} as a reference baseline. No human intervention (i.e. additional post processing) was applied to these outputs. 

\begin{table}[t]
\centering
\small
\setlength{\tabcolsep}{3mm}
\begin{tabular}{lc}
\toprule
\textbf{Dataset} & \textbf{Standard Error Rate} \\
\midrule
TacoPlay      & $0.01 \pm 0.01$ \\
RT-1          & $0.00 \pm 0.02$ \\
SCOUT         & $0.01 \pm 0.014$ \\
ALFRED        & $0.01 \pm 0.01$ \\
BRIDGE        & $0.02 \pm 0.02$ \\
LanguageTable & $0.01 \pm 0.01$ \\
LIBERO        & 0.01 (full dataset) \\
\bottomrule
\end{tabular}
\caption{Standard error rates across datasets. Values are reported as mean $\pm$ 95\% confidence interval with $n=200$, unless otherwise noted.}
\label{tab:pos_standard_error_rates}
\end{table}

\section{Verb, Direct Object, Adverbial Diversity.}
\label{do-method}

\begin{figure}
    \centering
    \begin{subfigure}{\linewidth}
        \centering
        \includegraphics[width=\linewidth]{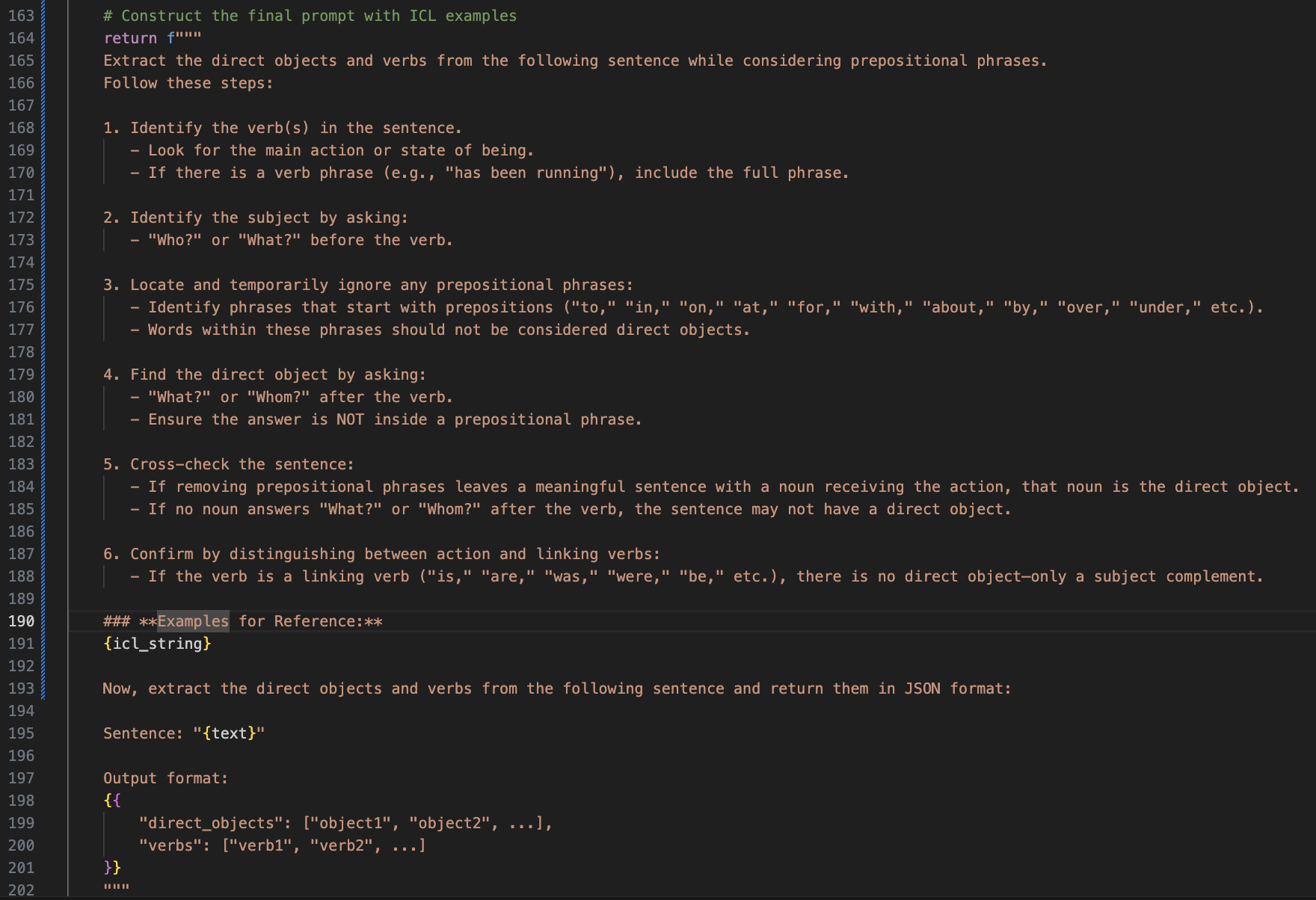}
        \caption{Verb–direct object prompt example used in the \textbf{``Verb, Direct Object, Adverbial Diversity"} section.}
        \label{fig:dir-obj-prompt}
    \end{subfigure}

    \begin{subfigure}{\linewidth}
        \centering
        \includegraphics[width=\linewidth]{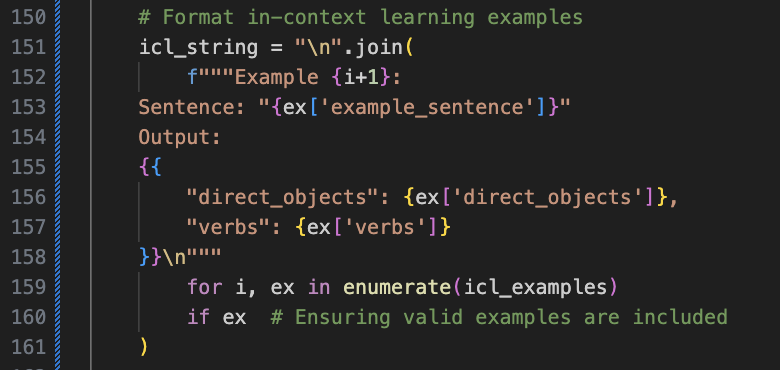}
        \caption{In context learning string generated by TF-IDF distance k-nearest neighbors.}
        \label{fig:icl-string}
    \end{subfigure}

    \caption{Prompts used in direct object and verb parsing tasks for instruction analysis.}
    \label{fig:dir-obj-parse-prompt}
\end{figure}

\begin{figure*}
    \centering
    \includegraphics[width=\linewidth]{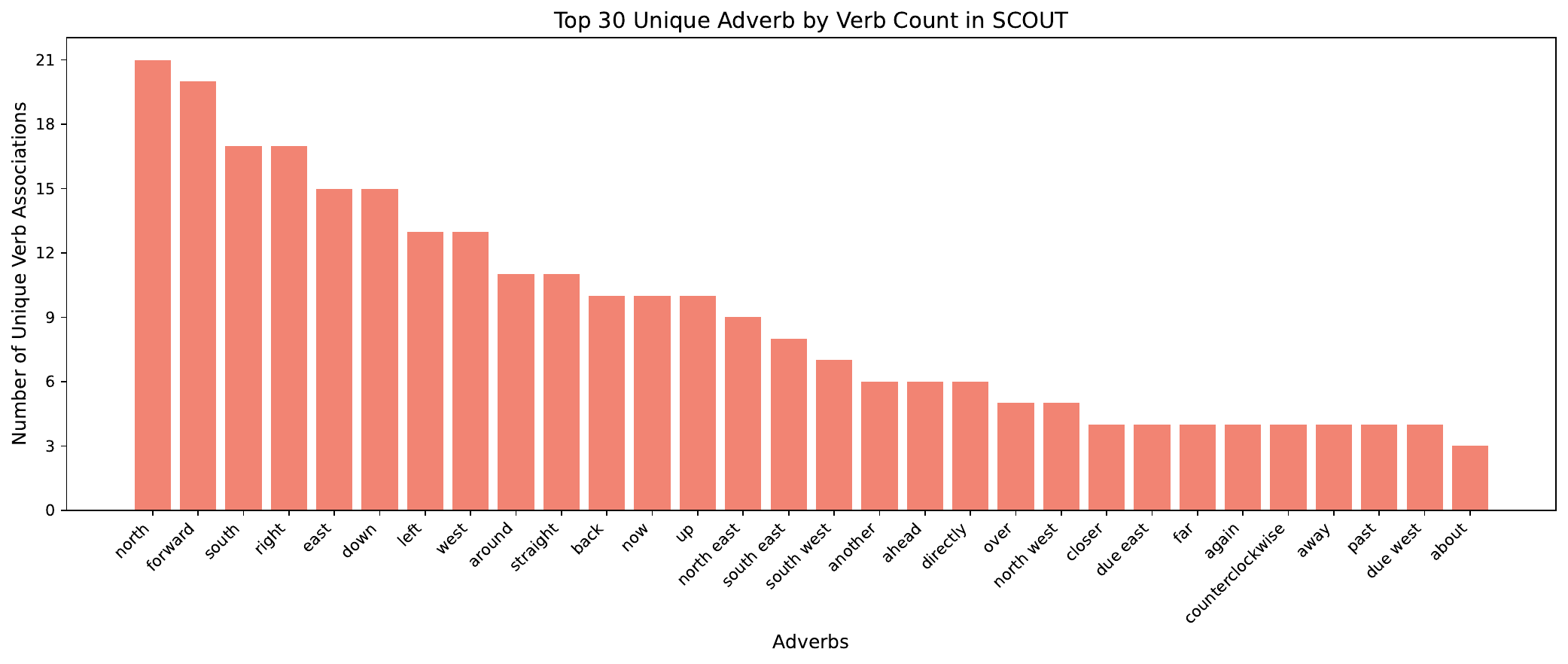}
    \caption{\textbf{Analysis 2: Semantic Diversity}  VLN adverbials - limited to the top 30 adverbs with most unique language use}
    \label{fig:scout-adverbial}
\end{figure*}

\begin{figure*}
    \centering
    \includegraphics[width=\linewidth]{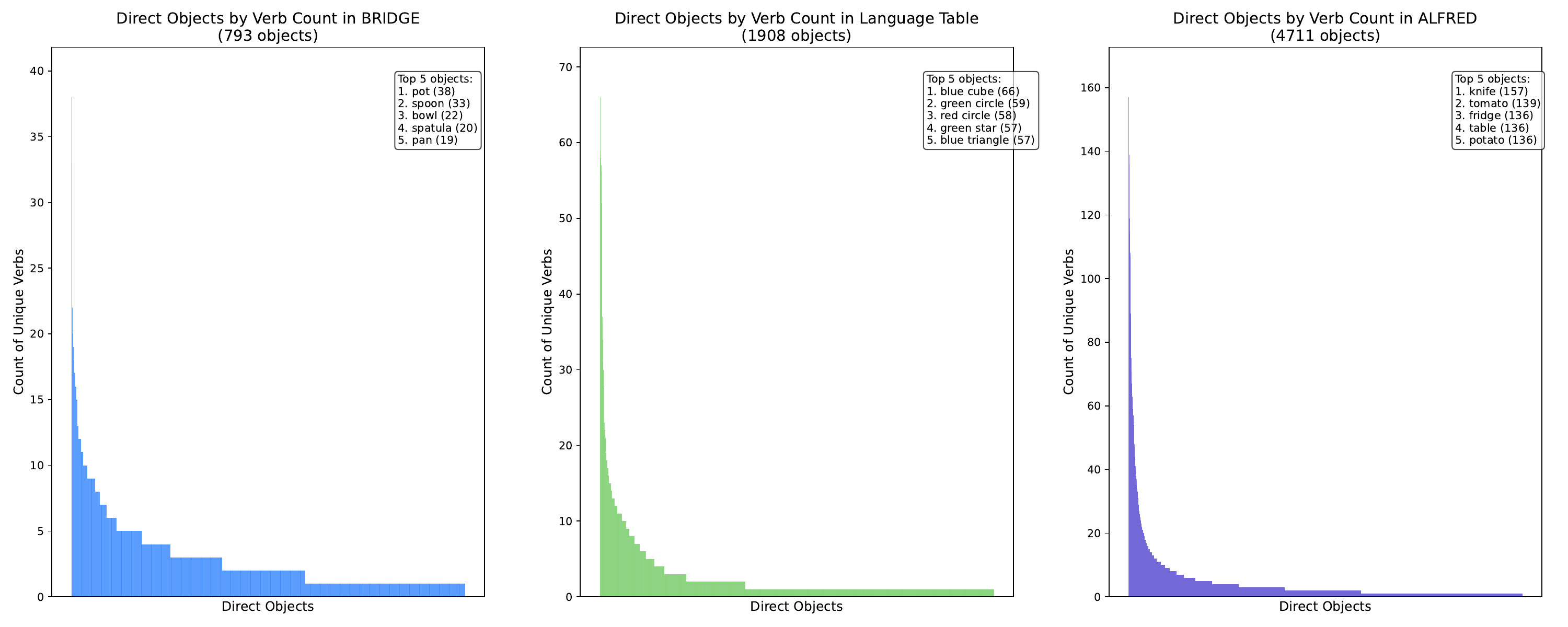}
    \caption{\textbf{Analysis 2: Semantic Diversity}  Frequency Plot of Unique Verbs per Direct Object for Manipulation Datasets}
    \label{fig:eai-dir-obj-histos-full}
\end{figure*}

\begin{figure*}
    \centering
    \includegraphics[width=0.8\linewidth]{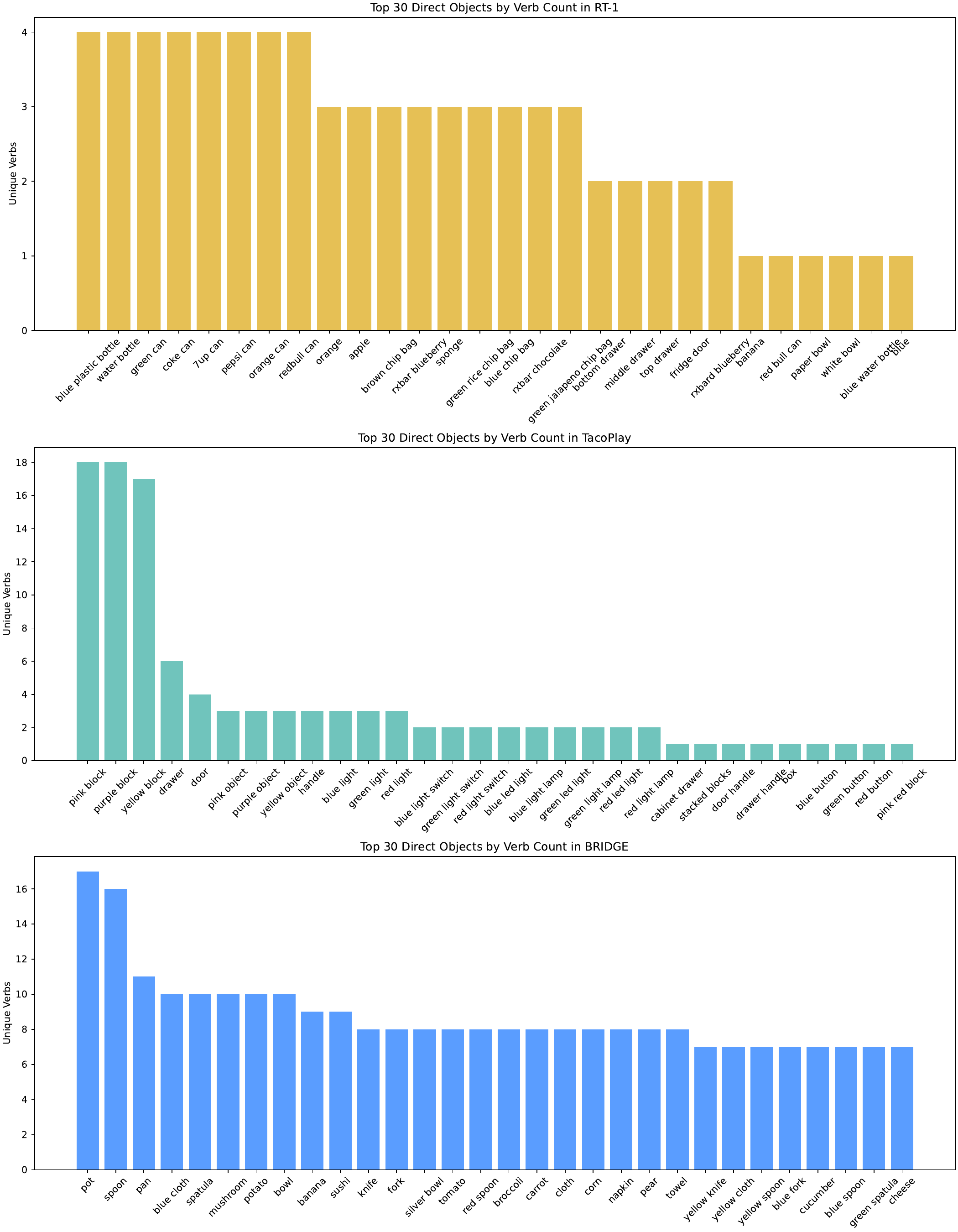}
    \caption{\textbf{Analysis 2: Semantic Diversity} Frequency Plot of Unique Verbs per Direct Object for Manipulation Datasets}
    \label{fig:eai-dir-obj-histos}
\end{figure*}

To extract verb, direct object, and adverbial features, an author manually annotated the LIBERO, TacoPlay, RT-1, SCOUT datasets, and Bridge datasets. For larger datasets: Language Table and ALFRED we implemented a large-scale annotation pipeline using \texttt{R1-Distill-Qwen-14B} \cite{deepseekai2025deepseekr1}. The model was loaded in 8-bit quantized format to optimize memory usage, and batch \(b = 10\) processing was employed to maximize throughput. The prompts for the model followed the format shown in Figure \ref{fig:dir-obj-parse-prompt}. We implemented in-context learning (ICL) on Language Table to enhance accuracy by retrieving sentence-specific examples using TF-IDF similarity. Despite using LLMs, all annotations were manually reviewed to ensure consistency, including lemmatizing verbs, removing duplicates, and normalizing synonymous expressions (e.g., ``pick" vs. ``pick up"). This hybrid method enabled the construction of high-quality annotations for downstream analysis. Results are provided in Figures \ref{fig:eai-dir-obj-histos}, \ref{fig:eai-dir-obj-histos-full}, and \ref{fig:scout-adverbial}.

\textbf{On Object and Adverbial Diversity.} We assessed how many distinct verbs are used with each direct object for manipulation datasets. Low counts suggest limited interaction diversity, sometimes due to real-world constraints, but often due to overly templated instruction generation. Direct object structures are less relevant for navigation-focused datasets, instead how an instruction is followed, e.g., directional terms (e.g., ``north," ``forward"), location-based modifiers (e.g., ``around," ``inside"), manner descriptors (e.g., ``slowly," ``directly") are more relevant.

\begin{figure*}
    \centering
        \centering
            \includegraphics[width=\textwidth]{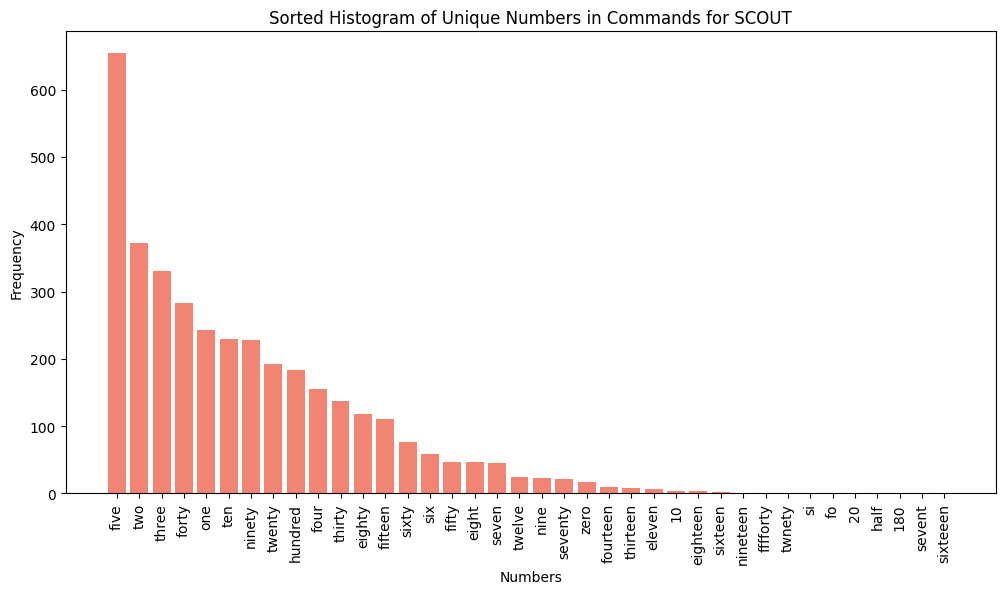}
            \caption{\textbf{Analysis 2: Semantic Diversity}  SCOUT Numerics}
            \label{fig:scout-numerics}
\end{figure*}

\begin{figure*}
        \centering
            \includegraphics[width=\textwidth]{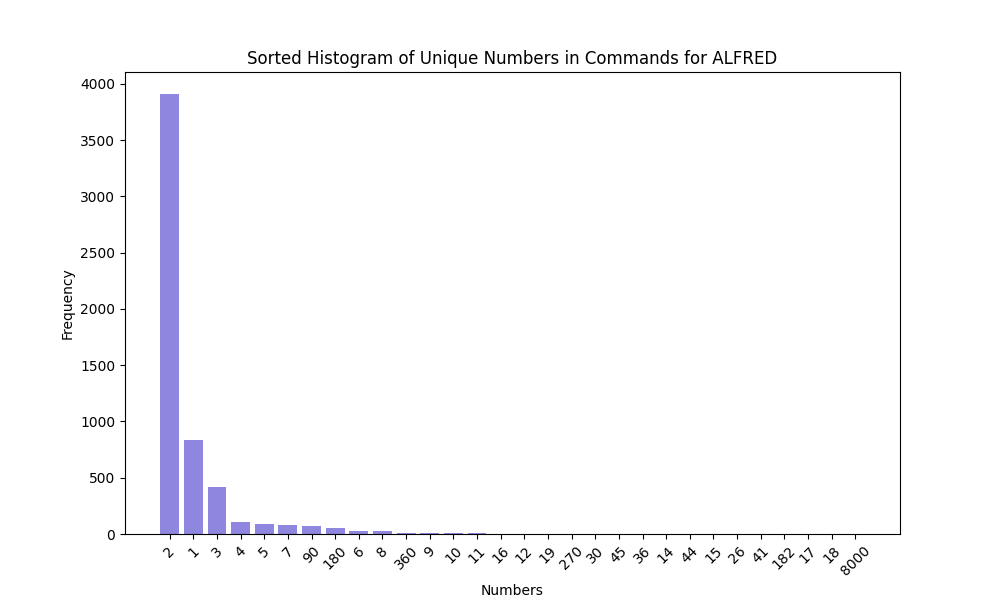}
            \caption{\textbf{Analysis 2: Semantic Diversity}  ALFRED Numerics}
            \label{fig:alfred-numerics}
    \caption{Numeric representation in navigation datasets.}
    \label{fig:nav-num}
\end{figure*}

\textbf{On Numeric Generalization.}  
As VLA models are increasingly expected to interpret numerical quantities (e.g., distances, angles) in an end-to-end manner, the distribution of numerical values in navigation instructions becomes more critical. Figure~\ref{fig:scout-numerics} shows that numbers like ``two," ``three," and ``five" are relatively common in SCOUT, while values such as ``seven," ``eight," or ``twelve" are rare. ALFRED (see Figure \ref{fig:alfred-numerics}) appears more tailed and its numeric coverage is weaker than SCOUT; however, the overall representation of numerics is greater due to dataset size. This sparsity raises concerns about whether models trained on these datasets can interpolate or generalize to underrepresented numerical instructions. For example, can a robot correctly interpret ``move seven meters" if it has never encountered that number in training? What if it has only encountered meters but is given a command in yards? What if the command contains common shortcuts, such as using 4K to refer to 4,000? Future research should investigate the impact of numeric and unit sparsity on navigation performance and explore methods for balancing numerical distributions during data collection or augmentation.

\textbf{On Annotation Quality.} For the dependency parsing results focused on direct object and adverbial analysis, we employed a multi-stage human-in-the-loop annotation pipeline. Initial verb–direct object pair predictions follow from our aforementioned procedure, as a coarse draft. This first-stage output was evaluated in a binary ``good''/``bad'' manner based on whether the correct verb–object pairs were present or if extra incorrect pairs were included. These predictions were not used as final outputs. The annotations underwent 2-3 rounds of cleaning, standardization, and corrections before being reported. We provide quantitative details on the initial draft error rates in Table \ref{tab:stage1_error_rates}, but we emphasize that these first-round predictions were not used in generating the final results. 

\begin{table*}[h]
\centering
\small
\setlength{\tabcolsep}{3mm}
\begin{tabular}{lcl}
\toprule
\textbf{Dataset} & \textbf{Stage 1 LLM Standard Error Rate} & \textbf{Final Stage} \\
\midrule
TacoPlay      & 0.34 (full dataset) & Manually annotated full dataset \\
RT-1          & 0.23 (full dataset) & Manually annotated full dataset \\
SCOUT         & --                  & Manually annotated full dataset \\
ALFRED        & $0.07 \pm 0.00$ (n=101, 95\% CI) & Manually annotated full dataset \\
BRIDGE        & $0.28 \pm 0.04$ (n=590, 95\% CI) & Manually annotated full dataset \\
LanguageTable & $0.08 \pm 0.04$ (n=200, 95\% CI) & Manually annotated full dataset \\
LIBERO        & $0.26 \pm 0.09$ (n=101, 95\% CI) & Manually annotated full dataset \\
\bottomrule
\end{tabular}
\caption{Stage 1 LLM standard error rates and final annotation stage across datasets. Error rates are reported either on the full dataset or as mean $\pm$ confidence interval (95\%) with sample size $n$.}
\label{tab:stage1_error_rates}
\end{table*}

\section{Instruction Structure Analysis}
\label{sec:appendix-instruction-structure}
To analyze the compositional structure of language in robotics datasets, we use LLM-generated feature information (see Appendices~\ref{pos-method} and~\ref{do-method}) to construct heuristics for detecting four types of instruction-level patterns: negation, conditionality, multi-step sequencing, and cyclical structures. These patterns are identified through string-matching techniques and syntactic cues extracted from dependency parses and part-of-speech tags.

\begin{table}[t!]
\begin{center}
\scriptsize
\setlength{\tabcolsep}{0.5mm}
\begin{tabular}{lcccc}
\toprule
Dataset & Negation & Conditional & Multi Step & Cycle\\
\midrule

\multicolumn{5}{l}{\textit{Language-Focused Robotics Datasets}} \\
\hspace{1mm}ALFRED \cite{Shridhar_2020_CVPR} & 22 & 275 & 56026 & 3313 \\
\hspace{1mm}SCOUT \cite{lukin-etal-2024-scout-situated} & 122 & 85 & 1890 & 379 \\
\midrule
\multicolumn{5}{l}{\textit{VLA Datasets}} \\
\hspace{1mm}RT-1 \cite{brohan2023rt1roboticstransformerrealworld} & 0 & 0 & 82 & 0\\
\hspace{1mm}BRIDGE \cite{walke2023bridgedata} & 27 & 2 & 3113 & 139 \\
\hspace{1mm}TacoPlay \cite{rosete2022tacorl} & 0 & 0 & 104 & 0 \\
\hspace{1mm}Language Table \cite{langtable} & 26 & 6 & 22164 & 4579 \\
\hspace{1mm}LIBERO \cite{liu2023liberobenchmarkingknowledgetransfer} & 0 & 0 & 959 & 0 \\
\bottomrule
\end{tabular}
\end{center}
\caption{\textbf{Analysis 3: Structural Diversity }Further details on the Instruction Structure Analysis. Raw counts corresponding to Figure~\ref{fig:instruct-struct}.}
\label{tab-non-seq}
\end{table}

\begin{itemize}
    \item \textbf{Negation} was detected using syntactic cues like neg dependencies and lexical markers (e.g., ``not”, ``don’t”, ``never”).

    \item \textbf{Conditionality} was identified via subordinating conjunctions (e.g., ``if”, ``unless”) and dependency markers indicating conditional clauses.
    
    \item \textbf{Multi-step} sequencing was inferred from coordinating conjunctions (e.g., ``and”, ``then”), punctuation, or imperative chaining.
    
    \item \textbf{Cyclical} patterns were identified using repeat verbs (``again”, ``repeat”) or constructions indicating iteration or loops.
\end{itemize}

For each instruction, we annotated binary indicators for each structure type and aggregated them to compute relative frequencies across datasets. Quantitative results are presented in Figure~\ref{fig:instruct-struct}, and representative examples are shown in Table~\ref{tab:instruction-examples}. These results help reveal structural tendencies in instruction design; particularly, the dominance of linear, stepwise instruction formats and the underrepresentation of more complex, logic-driven patterns.

\textbf{On Annotation Quality.} In Figure \ref{fig:instruct-struct}, for datasets with fewer than 600 unique instructions, we manually annotated all examples and computed inter-annotator agreement (IAA) between two annotators. Human–human annotations consisted of binary judgments (``yes''/``no'') indicating whether the LLM-generated annotation was correct, independent of the specific structural category predicted. Inter-annotator agreement was then computed to assess consistency in these correctness judgments. For the smaller datasets (RT-1, TacoPlay, and LIBERO), annotators achieved perfect agreement, with Cohen’s $\kappa = 1.0$. This may be due to the templated construction of the instructions for these datasets.

For datasets with more than 600 unique instructions, we randomly sampled 500 unique commands for manual annotation. We computed human agreement on this subset and used the same subset to estimate standard errors. To compute LLM–human agreement (IAA H-L) and the associated standard errors (SE\_{CATEGORY}), one annotator additionally recorded category-specific correctness (see Table \ref{tab:fig_4_iaa_se_results}). The low agreement in multi-step was due to a systematic ambiguity regarding whether compound objects (e.g., 'move X and Y') should be labeled by their syntax (single-step) or their robotic execution (multi-step).

\begin{table*}[h]
\begin{center}
\scriptsize
\setlength{\tabcolsep}{2.5mm}
\begin{tabular}{lccccccccc}
\toprule
Dataset & IAA H-H & \multicolumn{2}{c}{Negation} & \multicolumn{2}{c}{Conditional} & \multicolumn{2}{c}{Multi-Step} & \multicolumn{2}{c}{Cycle} \\
\addlinespace[0.25em]
\cmidrule(lr){3-4}\cmidrule(lr){5-6}\cmidrule(lr){7-8}\cmidrule(lr){9-10}
\addlinespace[0.25em]
 &  & SE & IAA H-L & SE & IAA H-L & SE & IAA H-L & SE & IAA H-L \\
\midrule
\hspace{3mm}ALFRED
& 0.66
& 0.00 & 1.00
& 0.00 & 1.00
& $0.04 \pm 0.02$ & 0.92
& $0.01 \pm 0.01$ & 0.87 \\

\hspace{3mm}SCOUT
& 0.88
& 0.00 & 1.00
& 0.00 & 1.00
& $0.17 \pm 0.03$ & 0.20
& $0.00 \pm 0.01$ & 0.95 \\

\hspace{3mm}BRIDGE
& 0.99
& $0.00 \pm 0.01$ & 0.50
& 0.00 & 1.00
& $0.10 \pm 0.03$ & 0.72
& $0.01 \pm 0.01$ & 0.66 \\

\hspace{3mm}LanguageTable
& 0.55
& 0.00 & 1.00
& 0.00 & 1.00
& $0.14 \pm 0.03$ & 0.29
& $0.01 \pm 0.01$ & 0.90 \\

\bottomrule
\end{tabular}
\end{center}
\caption{For datasets with more than 600 unique instructions, we randomly sampled 500 unique commands for manual annotation and calculated the following metrics. Inter-annotator agreement (IAA) as Cohen's $\kappa$ and Standard Errors (SE) are reported. Human-Human IAA is denoted by H-H. Human-LLM IAA is denoted by H-L.}
\label{tab:fig_4_iaa_se_results}
\end{table*}

\begin{figure*}[hp!]
\begin{center}
\small
\setlength{\tabcolsep}{3mm}
\renewcommand{\arraystretch}{1.6}
\begin{tabular}{@{}p{2cm} p{6cm} p{6cm}@{}}
\toprule
\textbf{Dataset} & \textbf{POS Pattern} & \textbf{Example Sentences} \\
\midrule

\multicolumn{3}{l}{\textit{TacoPlay}} \\[1mm]

 & VERB $\rightarrow$ DET $\rightarrow$ ADJ $\rightarrow$ NOUN $\rightarrow$ ADP $\rightarrow$ DET $\rightarrow$ NOUN 
& put the purple block on the table\newline slide the purple block to the left\newline place the yellow block on the table \\
\cmidrule(lr){2-3}

 & VERB $\rightarrow$ DET $\rightarrow$ ADJ $\rightarrow$ NOUN $\rightarrow$ ADP $\rightarrow$ DET $\rightarrow$ ADJ $\rightarrow$ NOUN 
& put the pink object inside the left cabinet\newline put the yellow block inside the right cabinet\newline place the purple block inside the right cabinet \\
\cmidrule(lr){2-3}

 & VERB $\rightarrow$ DET $\rightarrow$ ADJ $\rightarrow$ NOUN $\rightarrow$ CCONJ $\rightarrow$ VERB $\rightarrow$ PRON $\rightarrow$ ADV 
& take the purple block and rotate it right\newline take the yellow block and turn it right\newline grasp the purple block and turn it left \\
\midrule

\multicolumn{3}{l}{\textit{RT-1}} \\[1mm]

 & VERB $\rightarrow$ NOUN $\rightarrow$ NOUN $\rightarrow$ ADP $\rightarrow$ ADJ $\rightarrow$ NOUN 
& place rxbar blueberry into bottom drawer\newline move rxbar chocolate near orange can\newline move 7up can near green can \\
\cmidrule(lr){2-3}

 & VERB $\rightarrow$ NOUN $\rightarrow$ NOUN $\rightarrow$ ADP $\rightarrow$ NOUN $\rightarrow$ NOUN 
& move water bottle near rxbar chocolate\newline move coke can near water bottle\newline move rxbar blueberry near water bottle \\
\cmidrule(lr){2-3}

 & VERB $\rightarrow$ NOUN $\rightarrow$ NOUN $\rightarrow$ ADP $\rightarrow$ ADJ $\rightarrow$ NOUN $\rightarrow$ CCONJ $\rightarrow$ VERB $\rightarrow$ ADP $\rightarrow$ NOUN 
& pick coke can from bottom drawer and place on counter\newline pick water bottle from top drawer and place on counter\newline pick rxbar blueberry from middle drawer and place on counter \\
\bottomrule
\end{tabular}
\caption{\textbf{Analysis 3: Structural Diversity} POS patterns and example sentences from TacoPlay and RT-1. Each example sentence is aligned with its corresponding POS pattern, and grouped by dataset.}
\label{tab:pos-patterns-examples1}
\end{center}
\end{figure*}

\begin{figure*}[hp!]
\begin{center}
\small
\setlength{\tabcolsep}{3mm}
\renewcommand{\arraystretch}{1.6}
\begin{tabular}{@{}p{2cm} p{6cm} p{6cm}@{}}
\toprule
\textbf{Dataset} & \textbf{POS Pattern} & \textbf{Example Sentences} \\
\midrule
\multicolumn{3}{l}{\textit{SCOUT}} \\[1mm]

 & VERB $\rightarrow$ ADV $\rightarrow$ NUM $\rightarrow$ NOUN 
& turn left thirty degrees\newline turn left ninety degrees\newline move forward one foot \\
\cmidrule(lr){2-3}

 & VERB $\rightarrow$ ADP $\rightarrow$ DET $\rightarrow$ NOUN 
& move towards a shoe\newline move towards the barrel\newline go through the door \\
\cmidrule(lr){2-3}

 & VERB $\rightarrow$ NUM $\rightarrow$ NOUN $\rightarrow$ ADV 
& turn sixty degrees left\newline move ten inches northeast\newline move two feet forward \\
\midrule

\multicolumn{3}{l}{\textit{BRIDGE}} \\[1mm]

 & VERB $\rightarrow$ DET $\rightarrow$ NOUN $\rightarrow$ ADP $\rightarrow$ DET $\rightarrow$ NOUN $\rightarrow$ PUNCT 
& Place the mushroom behind the spatula.\newline Place the salmon in the pot.\newline Move the mushroom onto the towel. \\
\cmidrule(lr){2-3}

 & VERB $\rightarrow$ DET $\rightarrow$ NOUN $\rightarrow$ ADP $\rightarrow$ DET $\rightarrow$ NOUN $\rightarrow$ ADP $\rightarrow$ DET $\rightarrow$ NOUN $\rightarrow$ PUNCT 
& Move the spatula at the edge of the table.\newline Move the spoon to the left of the napkin.\newline Put the cloth to the left of the spoon. \\
\cmidrule(lr){2-3}

 & VERB $\rightarrow$ DET $\rightarrow$ NOUN $\rightarrow$ ADP $\rightarrow$ DET $\rightarrow$ ADJ $\rightarrow$ NOUN $\rightarrow$ PUNCT 
& Place the strawberry in the silver pot.\newline Set the pot onto the green cloth.\newline Place the pot on the blue cloth. \\
\bottomrule
\end{tabular}
\caption{\textbf{Analysis 3: Structural Diversity} POS patterns and example sentences from SCOUT and BRIDGE datasets. Each example sentence is aligned with its corresponding POS pattern, and grouped by dataset.}
\label{tab:pos-patterns-examples2}
\end{center}
\end{figure*}


\begin{figure*}[t]
\centering

\begin{tikzpicture}
\begin{axis}[
    xbar stacked,
    width=0.88\textwidth,
    height=0.60\textheight,
    xmin=0,
    xlabel={Frequency},
    ylabel={POS Pattern Index},
    title={Top 35 POS Patterns Across Datasets},
    y dir=reverse,
    ymin=0.5,
    ymax=35.5,
    ytick={1,...,35},
    yticklabel style={font=\normalsize},
    xticklabel style={font=\small},
    xlabel style={font=\small},
    ylabel style={font=\small},
    title style={font=\normalsize},
    bar width=5pt,
    enlarge y limits=false,
    tick align=outside,
    axis lines*=left,
    legend style={
        at={(0.98,0.02)},
        anchor=south east,
        draw=none,
        fill=white,
        fill opacity=0.85,
        text opacity=1,
        font=\small
    },
    legend cell align=left,
]

\pgfplotstableread[col sep=comma]{figures/pos_pattern_fig_redux/acl_top35_pos_patterns.csv}\datatable

\addplot+[draw=none, fill={rgb,255:red,224; green,177; blue,43}]
table[x=RT-1, y=Index] \datatable;

\addplot+[draw=none, fill={rgb,255:red,239; green,102; blue,80}]
table[x=SCOUT, y=Index] \datatable;

\addplot+[draw=none, fill={rgb,255:red,114; green,104; blue,216}]
table[x=ALFRED, y=Index] \datatable;

\addplot+[draw=none, fill={rgb,255:red,49; green,133; blue,255}]
table[x=BRIDGE, y=Index] \datatable;

\addplot+[draw=none, fill={rgb,255:red,77; green,182; blue,172}]
table[x=TacoPlay, y=Index] \datatable;

\addplot+[draw=none, fill={rgb,255:red,139; green,212; blue,126}]
table[x={LanguageTable}, y=Index] \datatable;

\addplot+[draw=none, fill={rgb,255:red,233; green,78; blue,119}]
table[x=LIBERO, y=Index] \datatable;

\addlegendimage{area legend, draw=none, fill={rgb,255:red,224; green,177; blue,43}}
\addlegendentry{RT-1}
\addlegendimage{area legend, draw=none, fill={rgb,255:red,239; green,102; blue,80}}
\addlegendentry{SCOUT}
\addlegendimage{area legend, draw=none, fill={rgb,255:red,114; green,104; blue,216}}
\addlegendentry{ALFRED}
\addlegendimage{area legend, draw=none, fill={rgb,255:red,49; green,133; blue,255}}
\addlegendentry{BRIDGE}
\addlegendimage{area legend, draw=none, fill={rgb,255:red,77; green,182; blue,172}}
\addlegendentry{TacoPlay}
\addlegendimage{area legend, draw=none, fill={rgb,255:red,139; green,212; blue,126}}
\addlegendentry{Language Table}
\addlegendimage{area legend, draw=none, fill={rgb,255:red,233; green,78; blue,119}}
\addlegendentry{LIBERO}

\end{axis}
\end{tikzpicture}

\vspace{0.3em}
\vspace{0.5em}

\noindent\textbf{POS Pattern Key:}

\vspace{0.5em}

\tiny
\renewcommand{\arraystretch}{1.05}

\begin{minipage}[t]{0.48\textwidth}
\begin{tabular}{r p{0.82\linewidth}}
\begin{filecontents*}{figures/pos_pattern_fig_redux/acl_top35_pos_pattern_key.csv}
Index,Pattern
1,VERB $\rightarrow$ DET $\rightarrow$ ADJ $\rightarrow$ NOUN $\rightarrow$ ADV $\rightarrow$ ADP $\rightarrow$ DET $\rightarrow$ ADJ $\rightarrow$ NOUN
2,VERB $\rightarrow$ DET $\rightarrow$ ADJ $\rightarrow$ NOUN $\rightarrow$ ADP $\rightarrow$ DET $\rightarrow$ ADJ $\rightarrow$ NOUN
3,VERB $\rightarrow$ DET $\rightarrow$ ADJ $\rightarrow$ NOUN $\rightarrow$ ADP $\rightarrow$ DET $\rightarrow$ NOUN $\rightarrow$ ADP $\rightarrow$ DET $\rightarrow$ ADJ $\rightarrow$ NOUN
4,VERB $\rightarrow$ DET $\rightarrow$ ADJ $\rightarrow$ NOUN $\rightarrow$ ADP $\rightarrow$ DET $\rightarrow$ ADJ $\rightarrow$ NOUN $\rightarrow$ ADP $\rightarrow$ DET $\rightarrow$ ADJ $\rightarrow$ NOUN
5,VERB $\rightarrow$ DET $\rightarrow$ NOUN $\rightarrow$ ADP $\rightarrow$ DET $\rightarrow$ NOUN $\rightarrow$ PUNCT
6,VERB $\rightarrow$ DET $\rightarrow$ NOUN $\rightarrow$ ADP $\rightarrow$ DET $\rightarrow$ NOUN
7,VERB $\rightarrow$ DET $\rightarrow$ ADJ $\rightarrow$ NOUN $\rightarrow$ ADP $\rightarrow$ DET $\rightarrow$ NOUN
8,VERB $\rightarrow$ ADJ $\rightarrow$ NOUN $\rightarrow$ ADP $\rightarrow$ ADJ $\rightarrow$ NOUN
9,VERB $\rightarrow$ DET $\rightarrow$ NOUN $\rightarrow$ ADP $\rightarrow$ DET $\rightarrow$ NOUN $\rightarrow$ ADP $\rightarrow$ DET $\rightarrow$ NOUN $\rightarrow$ PUNCT
10,VERB $\rightarrow$ DET $\rightarrow$ ADJ $\rightarrow$ NOUN $\rightarrow$ ADP $\rightarrow$ DET $\rightarrow$ NOUN $\rightarrow$ NOUN $\rightarrow$ ADP $\rightarrow$ DET $\rightarrow$ ADJ $\rightarrow$ NOUN
11,VERB $\rightarrow$ DET $\rightarrow$ ADJ $\rightarrow$ NOUN $\rightarrow$ ADP $\rightarrow$ DET $\rightarrow$ NOUN $\rightarrow$ PUNCT
12,VERB $\rightarrow$ ADP $\rightarrow$ DET $\rightarrow$ NOUN $\rightarrow$ ADP $\rightarrow$ DET $\rightarrow$ NOUN
13,VERB $\rightarrow$ DET $\rightarrow$ ADJ $\rightarrow$ NOUN $\rightarrow$ ADP $\rightarrow$ ADJ $\rightarrow$ NOUN
14,VERB $\rightarrow$ DET $\rightarrow$ ADJ $\rightarrow$ NOUN $\rightarrow$ ADP $\rightarrow$ DET $\rightarrow$ ADJ $\rightarrow$ ADJ $\rightarrow$ NOUN $\rightarrow$ ADP $\rightarrow$ DET $\rightarrow$ ADJ $\rightarrow$ NOUN
15,VERB $\rightarrow$ DET $\rightarrow$ ADJ $\rightarrow$ NOUN $\rightarrow$ ADJ $\rightarrow$ ADP $\rightarrow$ DET $\rightarrow$ ADJ $\rightarrow$ NOUN
16,VERB $\rightarrow$ ADP $\rightarrow$ DET $\rightarrow$ NOUN $\rightarrow$ ADP $\rightarrow$ DET $\rightarrow$ NOUN $\rightarrow$ PUNCT
17,VERB $\rightarrow$ DET $\rightarrow$ NOUN $\rightarrow$ ADP $\rightarrow$ DET $\rightarrow$ NOUN $\rightarrow$ ADP $\rightarrow$ DET $\rightarrow$ NOUN
18,VERB $\rightarrow$ ADP $\rightarrow$ DET $\rightarrow$ NOUN $\rightarrow$ ADP $\rightarrow$ DET $\rightarrow$ NOUN $\rightarrow$ ADP $\rightarrow$ DET $\rightarrow$ NOUN $\rightarrow$ PUNCT
19,VERB $\rightarrow$ ADP $\rightarrow$ DET $\rightarrow$ ADJ $\rightarrow$ NOUN $\rightarrow$ ADP $\rightarrow$ DET $\rightarrow$ NOUN $\rightarrow$ PUNCT
20,VERB $\rightarrow$ DET $\rightarrow$ NOUN $\rightarrow$ ADP $\rightarrow$ DET $\rightarrow$ ADJ $\rightarrow$ NOUN $\rightarrow$ PUNCT
21,VERB $\rightarrow$ ADV $\rightarrow$ NUM $\rightarrow$ NOUN
22,VERB $\rightarrow$ DET $\rightarrow$ ADJ $\rightarrow$ NOUN $\rightarrow$ ADP $\rightarrow$ DET $\rightarrow$ ADJ $\rightarrow$ NOUN $\rightarrow$ PUNCT
23,VERB $\rightarrow$ DET $\rightarrow$ NOUN $\rightarrow$ ADP $\rightarrow$ DET $\rightarrow$ ADJ $\rightarrow$ NOUN $\rightarrow$ ADP $\rightarrow$ DET $\rightarrow$ NOUN $\rightarrow$ PUNCT
24,VERB $\rightarrow$ DET $\rightarrow$ NOUN $\rightarrow$ ADP $\rightarrow$ DET $\rightarrow$ ADJ $\rightarrow$ NOUN
25,VERB $\rightarrow$ DET $\rightarrow$ NOUN $\rightarrow$ ADP $\rightarrow$ DET $\rightarrow$ ADJ $\rightarrow$ NOUN $\rightarrow$ ADP $\rightarrow$ DET $\rightarrow$ NOUN
26,VERB $\rightarrow$ ADP $\rightarrow$ DET $\rightarrow$ NOUN
27,VERB $\rightarrow$ DET $\rightarrow$ NOUN $\rightarrow$ ADP $\rightarrow$ DET $\rightarrow$ ADJ $\rightarrow$ ADJ $\rightarrow$ NOUN $\rightarrow$ ADP $\rightarrow$ DET $\rightarrow$ NOUN
28,VERB $\rightarrow$ NUM $\rightarrow$ NOUN $\rightarrow$ ADV
29,VERB $\rightarrow$ ADP $\rightarrow$ DET $\rightarrow$ ADJ $\rightarrow$ NOUN
30,VERB $\rightarrow$ ADV
31,VERB $\rightarrow$ NOUN
32,NUM $\rightarrow$ NOUN
33,VERB $\rightarrow$ NUM $\rightarrow$ NOUN
34,VERB $\rightarrow$ ADP $\rightarrow$ NOUN
35,VERB $\rightarrow$ NOUN $\rightarrow$ NOUN $\rightarrow$ ADP $\rightarrow$ ADJ $\rightarrow$ NOUN
\end{filecontents*}
\csvreader[
    head to column names,
    filter test={\ifnumless{\Index}{17}}
]{figures/pos_pattern_fig_redux/acl_top35_pos_pattern_key.csv}{}%
{\Index & \Pattern\\}
\end{tabular}
\end{minipage}
\hfill
\begin{minipage}[t]{0.48\textwidth}
\begin{tabular}{r p{0.82\linewidth}}
\csvreader[
    head to column names,
    filter test={\ifnumgreater{\Index}{16}}
]{figures/pos_pattern_fig_redux/acl_top35_pos_pattern_key.csv}{}%
{\Index & \Pattern\\}
\end{tabular}
\end{minipage}

\normalsize
\renewcommand{\arraystretch}{1.0}
\caption{Top 35 POS patterns across datasets after aggregating the top 10 POS patterns within each dataset.}
\label{fig:dep-parse-histo}
\end{figure*}


\newcommand{\pospanel}[3]{%
\begin{minipage}[t]{0.25\textwidth}
\centering
\begin{tikzpicture}
\begin{axis}[
    xbar,
    width=\textwidth,
    height=0.20\textheight,
    xmin=0,
    enlarge x limits={upper, value=0.08},
    y dir=reverse,
    ytick={1,2,3,4,5,6,7,8,9,10},
    yticklabels={1,2,3,4,5,6,7,8,9,10},
    yticklabel style={font=\tiny},
    xticklabel style={font=\scriptsize},
    title style={font=\small},
    xlabel style={font=\scriptsize},
    ylabel style={font=\scriptsize},
    tick align=outside,
    axis lines*=left,
    bar width=5pt,
    scale only axis,
    title={#2},
    xlabel={Percentage (\%)},
    ylabel={POS Pattern Index}
]
\addplot+[draw=none, fill=#3]
table[col sep=comma, x=Pct, y=Rank] {#1};
\end{axis}
\end{tikzpicture}

\vspace{0.2em}

\tiny
\begin{tabular}{r p{0.95\linewidth}}
\csvreader[
    head to column names
]{#1}{}%
{\Rank & \Pattern\\}
\end{tabular}
\end{minipage}
}

\begin{figure*}[t]
\centering

\pospanel{figures/pos_pattern_fig_redux/acl_fig2_BRIDGE.csv}{BRIDGE}{{rgb,255:red,49; green,133; blue,255}}
\hspace{0.05\textwidth}
\pospanel{figures/pos_pattern_fig_redux/acl_fig2_LanguageTable.csv}{LanguageTable}{{rgb,255:red,139; green,212; blue,126}}
\hspace{0.05\textwidth}
\pospanel{figures/pos_pattern_fig_redux/acl_fig2_ALFRED.csv}{ALFRED}{{rgb,255:red,114; green,104; blue,216}}

\vspace{1em}

\pospanel{figures/pos_pattern_fig_redux/acl_fig2_SCOUT.csv}{SCOUT}{{rgb,255:red,239; green,102; blue,80}}

\caption{Top 10 POS patterns by dataset (BRIDGE, LanguageTable, ALFRED, and SCOUT), normalized by dataset size. }
\label{fig:by-ds-dep-histo-grid-1}
\end{figure*}

\begin{figure*}[t]
\centering

\pospanel{figures/pos_pattern_fig_redux/acl_fig2_RT_1.csv}{RT-1}{{rgb,255:red,224; green,177; blue,43}}
\hspace{0.05\textwidth}
\pospanel{figures/pos_pattern_fig_redux/acl_fig2_LIBERO.csv}{LIBERO}{{rgb,255:red,233; green,78; blue,119}}
\hspace{0.05\textwidth}
\pospanel{figures/pos_pattern_fig_redux/acl_fig2_TacoPlay.csv}{TacoPlay}{{rgb,255:red,77; green,182; blue,172}}

\caption{Top 10 POS patterns by dataset (RT-1, LIBERO, TACOPLAY), normalized by dataset size. }
\label{fig:by-ds-dep-histo-grid-2}
\end{figure*}

\begin{figure*}
\centering
\includegraphics[width=\textwidth]{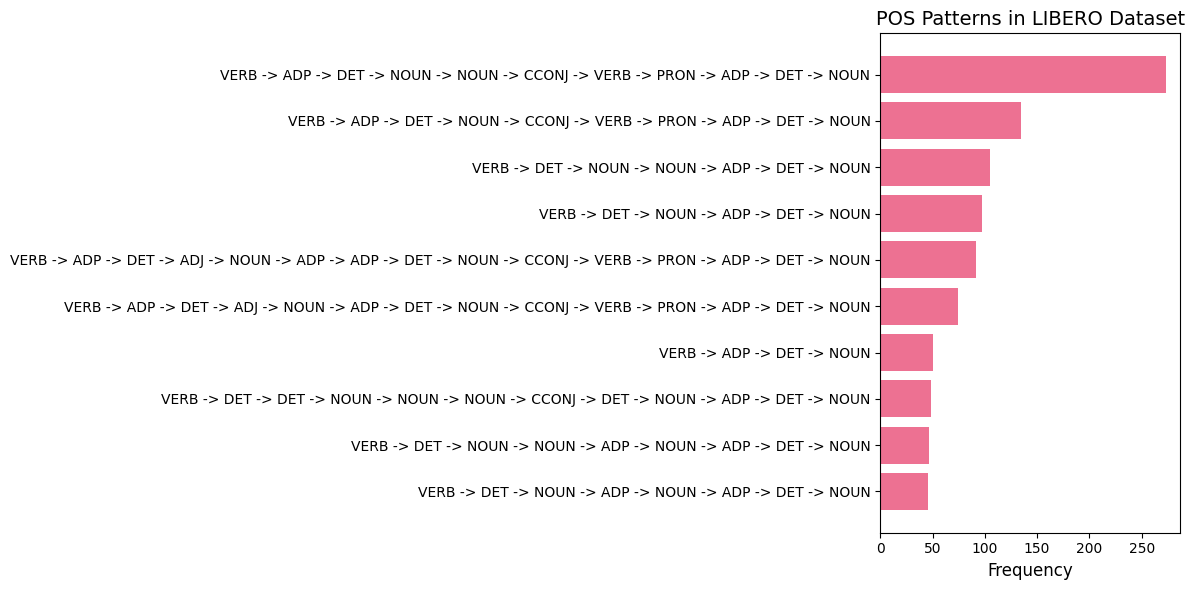}
\caption{\textbf{Analysis 3: Structural Diversity} Dependency parse features across \textbf{all} LIBERO splits.}
\label{fig:lib-dep-parse}
\end{figure*}

\begin{figure*}
\centering
\includegraphics[width=\textwidth]{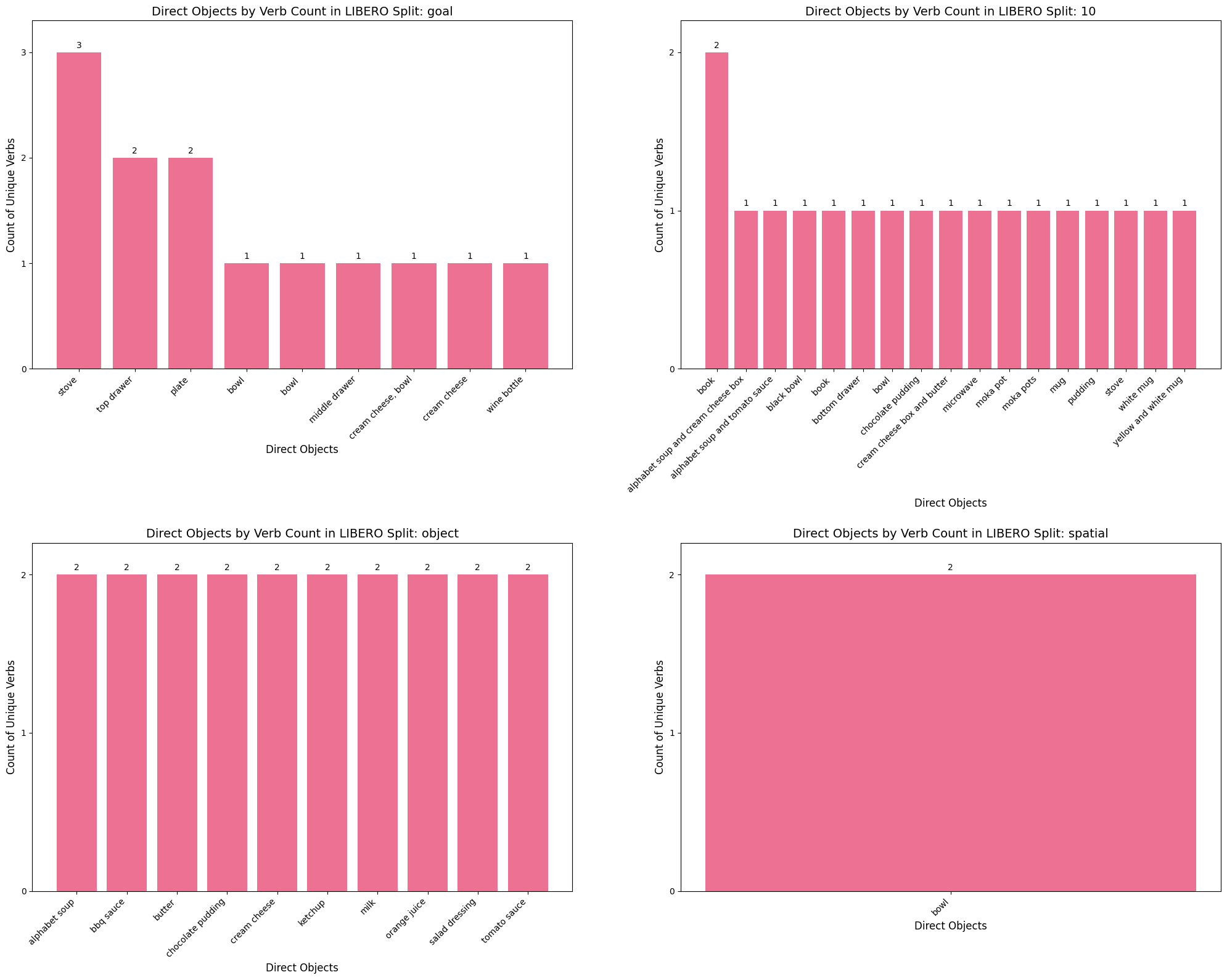}
\caption{\textbf{Analysis 2: Semantic Diversity} Verb and direct object frequencies across \textbf{all }LIBERO splits.}
\label{fig:lib-dir-obj}
\end{figure*}

\clearpage

\begin{table*}[t]
\centering
\small
\begin{tabularx}{\textwidth}{|l|l|X|}
\hline
\textbf{Category} & \textbf{Dataset} & \textbf{Examples} \\
\hline

\multirow{3}{*}{Negation} 
& SCOUT & i don't know what the red thing was \\
& & you are not at the total entrance \\
& & no i did not see any \\
\cline{2-3}
& BRIDGE & video frames not showing \\
& & video frames or not showing \\
& & Picture is not downloading, not able to view. \\
\cline{2-3}
& ALFRED & This step does not exist. \\
& & Slice the tomato on the counter but do not put down the knife. \\
& & Cook the potato slice in the microwave and do not put the cooked potato slice on the counter. \\
\hline

\multirow{3}{*}{Conditional} 
& SCOUT & see if there's a doorway \\
& & check and see if there's a doorway there \\
& & and i'll point out when there's a doorway so we can count them \\
\cline{2-3}
& BRIDGE & Pick the orange towel and place it on the middle if the table \\
& & PLACE THE YELLOW TOPWEL SIDE IF THE TABLE \\
\cline{2-3}
& ALFRED & Take keys from the black table, leave them on the lamp when you turn it on. \\
& & Turn right and walk until you're even with the fridge on your right and when you are turn right and walk to it. \\
& & Turn left and walk to the table then turn right when you get to it. \\
\hline

\multirow{6}{*}{Multi-Step} 
& LIBERO & open the top drawer and put the bowl inside \\
\cline{2-3}
& TacoPlay & go towards the drawer and place the pink object \\
& & go towards the purple block and grasp it \\
& & take the purple block and rotate it right \\
\cline{2-3}
& RT-1 & pick coke can from bottom drawer and place on counter \\
& & pick apple from top drawer and place on counter \\
& & pick green rice chip bag from bottom drawer and place on counter \\
\cline{2-3}
& SCOUT & and take a picture \\
& & and then the last question here anything that indicates the environment was recently occupied \\
& & and then take a picture \\
\cline{2-3}
& BRIDGE & put pot or pan on stove and put egg in pot or pan \\
& & Take the spatula from the vessel and place it on the table. \\
\cline{2-3}
& ALFRED & Open the drawer. Put the cell phone in the drawer on the right side towards the back and close it. \\
& & open the top right drawer of the desk, put phone inside, close the drawer \\
& & Turn and move to the far end of the kitchen island, so you're facing the tomato and fork. \\
\hline

\multirow{3}{*}{Cycle} 
& SCOUT & continue moving forward \\
& & follow hallway to the end of the wall uh to until you reach the wall \\
& & take a photo every forty five degrees \\
\cline{2-3}
& BRIDGE & end effector reaching knife \\
& & pick orange toy from vessel and keep it on the left side of the table \\
& & end effector reaching corn \\
\cline{2-3}
& ALFRED & Move over to the right side of the desk again. \\
& & Put the potato slice in the fridge and shut the door and then take the potato slice out and shut the fridge door again. \\
& & Walk to your left until you see a loaf of bread on the counter top. \\
\hline

\end{tabularx}
\caption{ \textbf{Analysis 3: Structural Diversity }
Representative instruction examples for negation, conditional, multi-step, and cycle structures. Note that in BRIDGE and ALFRED, some examples contain noise from the original OXE metadata (e.g., typos or syntactic errors); and in many cases, this noise artificially inflate diversity scores.
}
\label{tab:instruction-examples}
\end{table*}

\end{document}